\pdfoutput=1

\documentclass[11pt]{article}

\usepackage[]{ACL2023}

\usepackage{times}
\usepackage{latexsym}

\usepackage[T1]{fontenc}
\usepackage[export]{adjustbox}

\usepackage[utf8]{inputenc}

\usepackage{microtype}
\usepackage{algorithm}
\usepackage{algorithmic}
\usepackage{booktabs}
\usepackage{amsmath}
\usepackage{multirow}
\usepackage{bbding}
\usepackage{mathrsfs}
\usepackage{subfigure}
\usepackage{amssymb}
\usepackage{makecell}
\usepackage{graphicx}
\usepackage{inconsolata}

\title{FutureTOD: Teaching Future Knowledge to Pre-trained Language Model for Task-Oriented Dialogue}

\author{Weihao Zeng$^{1*}$, Keqing He$^{2*}$, Yejie Wang$^{1}$, Chen Zeng$^{1}$ \\ {\bf Jingang Wang$^{2}$,} {\bf Yunsen Xian$^{2}$,}  {\bf Weiran Xu$^{1}$}\thanks{\ \ The first two authors contribute equally. Weiran Xu is the corresponding author.}\\
  $^1$Beijing University of Posts and Telecommunications, Beijing, China\\
$^{2}$Meituan, Beijing, China\\
  \texttt{\{zengwh,wangyejie,chenzeng,xuweiran\}@bupt.edu.cn}\\
  \texttt{\{hekeqing,wangjingang,xianyunsen\}@meituan.com}
  }

\begin{document}
\maketitle
\begin{abstract}
Pre-trained language models based on general text enable huge success in the NLP scenario. But the intrinsical difference of linguistic patterns between general text and task-oriented dialogues makes existing pre-trained language models less useful in practice. Current dialogue pre-training methods rely on a contrastive framework and face the challenges of both selecting true positives and hard negatives. In this paper, we propose a novel dialogue pre-training model, FutureTOD, which distills future knowledge to the representation of the previous dialogue context using a self-training framework. Our intuition is that a good dialogue representation both learns local context information and predicts future information. Extensive experiments on diverse downstream dialogue tasks demonstrate the effectiveness of our model, especially the generalization, robustness, and learning discriminative dialogue representations capabilities. \footnote{Our code, models and other related resources are publicly available at \href{https://github.com/Zeng-WH/FutureTOD}{https://github.com/Zeng-WH/FutureTOD}}
\end{abstract}

\section{Introduction}

Pre-trained language models \cite{devlin-etal-2019-bert,Liu2019RoBERTaAR} based on a massive scale of general text corpora \cite{Zhu2015AligningBA} have been commonly used in many NLP applications. Finetuning models on these PLMs significantly improves the performance of various downstream tasks, especially natural language understanding. Despite their success, directly applying them to conversational corpora is proved to be suboptimal due to the large linguistic gap between conversations and plain text \cite{Rashkin2019TowardsEO,Wolf2019TransferTransfoAT}. Therefore, it's vital to explore dialogue-specific pre-trained models for solving various downstream dialogue tasks.

Early pre-trained dialogue language models use chit-chat corpora from social media, such as Twitter or Reddit, aiming at retrieval \cite{Henderson2019TrainingNR} and dialogue response generation \cite{Zhang2020DIALOGPTL}. These open-domain dialogues are usually short, noisy, and without specific chatting goals. Further, a more practical scenario, task-oriented dialogue (TOD), is attracting more attention. TOD has explicit goals (e.g. restaurant reservation) and many conversational interactions like belief states and database information, making language understanding and policy learning more complex than those chit-chat scenarios. Each TOD dataset is usually small because collecting and labeling such data are time-consuming. Therefore, in this paper, we focus on unsupervised dialogue pre-training for task-oriented dialogues.

\begin{figure}[t]
 \centering
 \setlength{
 \abovecaptionskip}{-0.1cm}
\resizebox{0.48\textwidth}{!}{
 \includegraphics[scale=0.7]{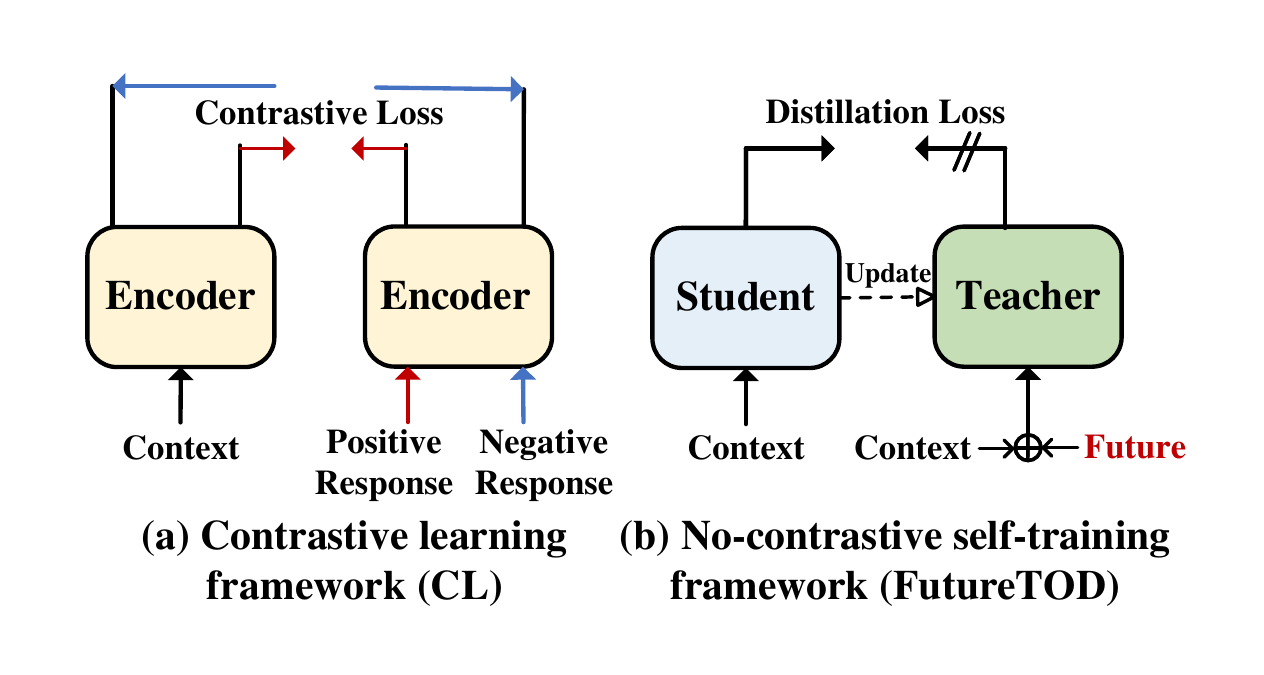}
 }

 \caption{Comparison of different dialogue pre-training paradigms. The contrastive models learn context representations by pulling together positive pairs and pushing apart negative pairs. In contrast, our FutureTOD employs a self-training framework to distill future knowledge to context representations and dismiss the requirements of contrastive pairs.}
 \label{fig:intro}

\end{figure} 

 \begin{figure*}[t]
 \centering
\resizebox{0.85\textwidth}{!}{
 \includegraphics[scale=0.5]{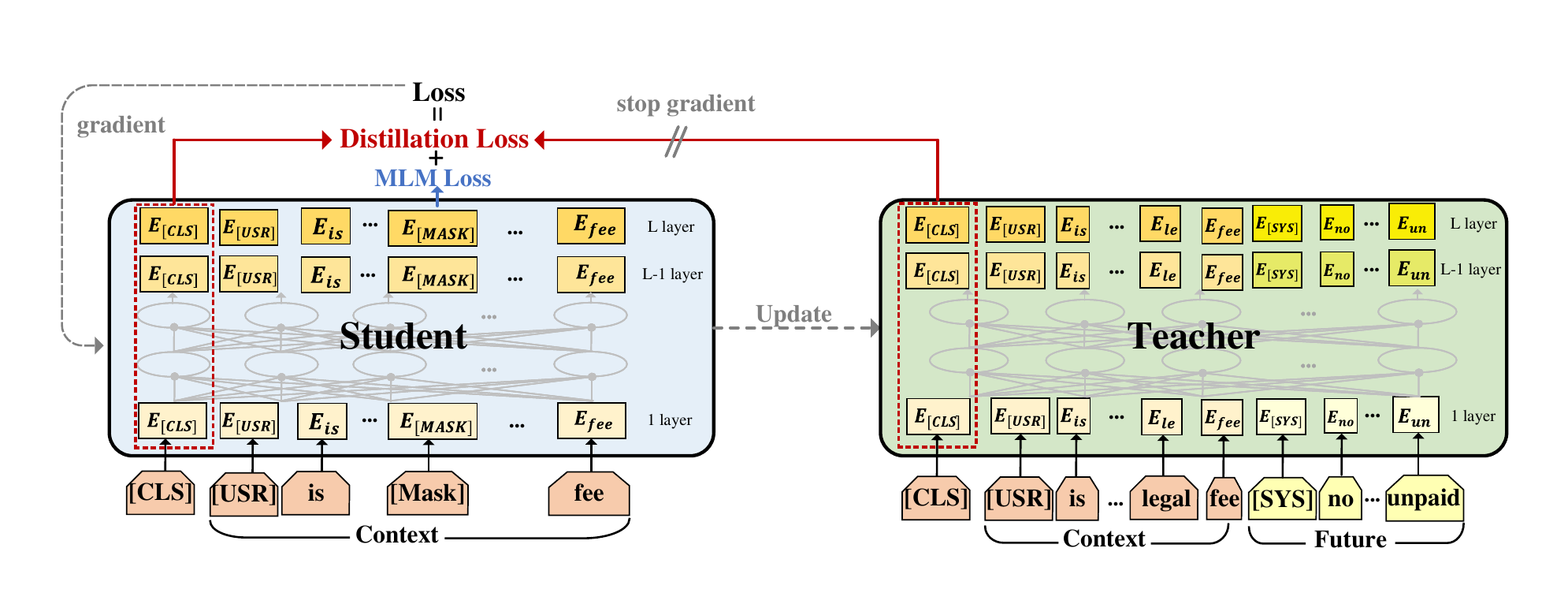}
 }
\vspace{-0.3cm}
 \caption{Overall architecture of FutureTOD. For brevity, we show only one system response utterance as future.}
 \label{fig:model}
 \vspace{-0.3cm}

\end{figure*} 

Previous TOD pre-training methods usually follow a contrastive learning (CL) framework \cite{chen2020simple,He2020MomentumCF} as shown in Figure \ref{fig:intro}(a). CL aims to pull together semantically similar (positive) pairs and push apart semantically dissimilar (negative) pairs. SimCSE \cite{gao-etal-2021-simcse} employs Dropout \cite{Srivastava2014DropoutAS} augmentation to construct positive pairs by passing a sentence through the encoder twice, resulting in superior performance for learning plain text representations. However, it performs poorly in the dialogue domain because of ignoring the intrinsic properties of dialogue data \cite{Zhou2022LearningDR}. TOD-BERT \cite{Wu2020TODBERTPN} takes the dialogue context\footnote{Throughout this paper, we denote a system turn including all the system sentences as the response (utterance), and all the history turns as the dialogue context.} and next response as a positive pair thus achieving promising performance on the response selection task. However, there is a large discrepancy in both semantics and data statistics between each response and its context \footnote{In the implementation of TOD-BERT, the context is often the concatenation of 5 to 15 utterances but the response is only a single utterance.}, which reduces its generalization ability to other dialogue tasks. Further, DSE \cite{Zhou2022LearningDR} learns from dialogues by taking consecutive utterances of the same dialogue as positive pairs. But the assumption that consecutive utterances represent similar semantics fails sometimes when answers are general and ubiquitous. Along with the issues of choosing positive pairs, these models regard other instances in the same batch as negative samples, which also induces potential noise to contrastive learning \cite{Arora2019ATA}, such as false negatives \cite{Huynh2022BoostingCS,Chen2022IncrementalFN} and relying on a large batch size \cite{He2020MomentumCF}. Overall, these contrastive methods face the challenges of both selecting true positive pairs and negative pairs that we aim to solve using a new non-contrastive pre-training framework.

In this paper, we propose a novel dialogue pre-training model, FutureTOD, which distills future knowledge to the representation of the previous dialogue context using future utterances based on a standard Transformer architecture BERT \cite{devlin-etal-2019-bert}. We argue that a good dialogue representation both learns local context information and predicts future knowledge. Instead of existing contrastive works, we employ a self-training framework and dismiss the requirements of contrastive pairs. As shown in Figure \ref{fig:intro}(b), we first use a student model to construct the dialogue representation of an input dialogue context. Next, we concatenate the context and following utterances and get its full representation using a teacher model. Our goal is to align the original context representation with the full representation containing future knowledge. The weights of the teacher are updated by the student periodically \cite{He2020MomentumCF,Baevski2022data2vecAG,Liu2022ExploringTR}. We evaluate FutureTOD on various task-oriented dialogue tasks, including intent classification, out-of-domain detection, dialogue state tracking, dialogue act prediction, and response selection. Experiment results demonstrate that FutureTOD significantly outperforms TOD-BERT, DSE, and other strong baselines in all the scenarios. We also observe FutureTOD has stronger capabilities on generalization, robustness and learning discriminative representations.

Our contributions are: (1) We propose a novel TOD dialogue pre-training model, FutureTOD, which distills future knowledge to dialogue representations. To the best of our knowledge, we are the first to use a non-contrastive self-training framework and knowledge distillation for dialogue pre-training. (2) Our model achieves consistent improvements on diverse downstream dialogue tasks over strong baselines. Extensive analyses prove the generalization, robustness, and learning discriminative dialogue representations capabilities.

\section{Model}
\subsection{Overall Architecture}
The overall architecture of FutureTOD is shown in Figure \ref{fig:model}. We adopt BERT-base-uncased\footnote{https://huggingface.co/bert-base-uncased} as our backbone following TOD-BERT \cite{Wu2020TODBERTPN}. We first add two special role tokens [USR] or [SYS] to the prefix of each utterance and concatenate all the utterances in the same dialogue into one flat sequence. Then we split each dialogue at a randomly selected turn t to get the context and future sequences. We encode the context using a student model and obtain the output of [CLS] as the original dialogue representation. Next, we construct training targets by encoding the context and future using a teacher model. Both the student and teacher are the same BERT but the weights of the teacher are updated by the student periodically. The learning goal is to align the original context representation with the full representation containing future knowledge. We assume a good dialogue representation can't only capture local context information but also predict future knowledge.

\subsection{Learning Future Knowledge}
\textbf{Notation} We use the collected datasets by TOD-BERT \cite{Wu2020TODBERTPN} as our pre-training corpus. 
For each dialogue, we first transform it into a token sequence. Following previous work \cite{Wu2020TODBERTPN,Zhou2022LearningDR}, we add two special role tokens [USR] or [SYS] to the prefix of each utterance and concatenate all the utterances into one flat sequence $D=\left\{U_{1}, S_{1}, \ldots, U_{n}, S_{n}\right\}$. $U_{1}$ and $S_{1}$ denotes the user utterance and system utterance, respectively. $n$ is the turn number of the dialogue.

\textbf{Learning Framework} Different from existing contrastive methods, we employ a self-training \cite{vanEngelen2019ASO,grill2020bootstrap} framework to distill future knowledge to the representation of the dialogue context using future utterances. The advantages are two-fold: (1) Our self-training framework doesn't require contrastive pairs thus alleviating the noise of selecting positive and negative samples. (2) Learning future knowledge encourages the model to align representations in the same latent space instead of pulling together representations of context and response belonging to different distributions. We first split each dialogue at a randomly selected turn t, so we get the context $C=\left\{U_{1}, S_{1}, \ldots, U_{t}\right\}$ and the future $F=\left\{S_{t}, U_{t+1}, S_{t+1}, \ldots, U_{n}, S_{n}\right\}$. Then we use a student model to encode the context and a teacher model to encode the context with the future. We denote the [CLS] output of the student model as $h_{S}$ and the teacher as $h_{T}$. We hope the student model can capture future information while modeling the local semantics. So we design a distillation loss $\mathcal{L}_{dis}$ by minimizing the discrepancy between $h_{S}$ and $h_{T}$:
{\setlength{\abovedisplayskip}{0.2cm}
\setlength{\belowdisplayskip}{0.2cm}
\begin{align}
   \mathcal{L}_{dis} = \left\|h_{S}-h_{T} \right\|_{2}
\end{align}
\label{max_future}
}
To explore different granularity of future information, we randomly select a ratio of future utterances from one utterance $S_{t}$ to the whole utterances $\left\{S_{t}, U_{t+1}, S_{t+1}, \ldots, U_{n}, S_{n}\right\}$. Besides, we find performing distillation loss on multiple layers rather than only the top layer also gives consistent improvements (see Section \ref{layer}). So, the final distillation loss $\mathcal{L}_{dis}$ is:
{\setlength{\abovedisplayskip}{0.1cm}
\setlength{\belowdisplayskip}{0.1cm}
\begin{align}\label{dis}
\mathcal{L}_{dis} = \sum_{l=1}^{L}(\left\|h_{S}^{l}-h_{T}^{l} \right\|_{2})
\end{align}
}
where $l$ is the $l$-th layer of BERT-base. We also try to apply normalization to $h_{S}$ and $h_{T}$ and other distillation objectives but do not observe significant change. Along with $\mathcal{L}_{dis}$, we also keep the traditional masked language modeling (MLM) \cite{devlin-etal-2019-bert} loss $L_{m l m}=-\sum_{m=1}^M \log P\left(x_m\right)$ following \citet{Wu2020TODBERTPN}, where $M$ is the total number of masked tokens and $P(x_{m})$ is the predicted probability of the token $x_{m}$ over the vocabulary size. Note that we only perform MLM on the student model. Therefore, the total loss is:
{\setlength{\abovedisplayskip}{0.1cm}
\setlength{\belowdisplayskip}{0.1cm}
\begin{align}\label{eq_dis}
\mathcal{L} = \mathcal{L}_{dis} + \mathcal{L}_{mlm}
\end{align}}
We simply sum them up and achieve the best performance in our experiments. 

\textbf{Parameter Updating} We employ a simple algorithm to optimize the parameters of the student and teacher models iteratively. (1) \textbf{Stage 1}: We first use Eq \ref{eq_dis} to perform gradient updating to optimize the student model and keep the teacher model fixed. We denote the interval as $E$ epochs.\footnote{We empirically find $E=10$ is the best. Please see a more detailed analysis in Section \ref{e}.} (2) \textbf{Stage 2}: After Stage 1, we directly assign student parameters to the teacher. The process of our method is summarized in Algorithm \ref{algo}.

\begin{algorithm}
	\renewcommand{\algorithmicrequire}{\textbf{Input:}}
	\renewcommand{\algorithmicensure}{\textbf{Output:}}
	\caption{FutureTOD}
	\label{alg1}
	\begin{algorithmic}[1]
		\STATE \textbf{Initialization:} Teacher $T$, Student $S$, Interval $E$, Total Epoch $M$
            \STATE \textbf{Input:} Context $C$, Future $F$
            \FOR{$m$ in [1, M]}
		\STATE Using $S$ to get the output $h_{S}$ of $C$
		\STATE Using $T$ to get the output $h_{T}$ of $C+F$
            \STATE Calculating the distillation loss $L_{dis}$ in Equation~\ref{dis}
		\STATE Calculating the MLM loss $L_{mlm}$
            \STATE Using $L=L_{dis}+L_{mlm}$ to update S
            \IF {$m$ \% $E$ == 0}
            
            \STATE Assigning $S$ parameters to the $T$
            \ENDIF
            \ENDFOR
		\ENSURE  $S$
	\end{algorithmic}  
 \label{algo}
\end{algorithm}

\section{Experiment}

\subsection{Pre-training Setting}
\textbf{Pre-training Corpus} We use the corpus collected by \citet{Wu2020TODBERTPN}, including 9 publicly available task-oriented datasets: MetaLWOZ \cite{Lee2019MultiDomainTD}, Schema \cite{Rastogi2020TowardsSM}, Taskmaster \cite{Byrne2019Taskmaster1TA}, MWOZ \cite{Budzianowski2018MultiWOZA}, MSR-E2E \cite{Li2018MicrosoftDC}, SMD \cite{Eric2017KeyValueRN}, Frames \cite{Asri2017FramesAC}, WOZ \cite{Mrksic2017NeuralBT}, CamRest676 \cite{RojasBarahona2017ANE}. We show the full statistics in Appendix \ref{data}.

\textbf{Baselines} We compare FutureTOD with other strong baselines. BERT \cite{devlin-etal-2019-bert} and BERT-mlm denotes the original BERT-base-uncased pre-trained on a large text corpus and continual pre-trained BERT using MLM on our pre-training dialogue corpus, respectively. DialoGPT \cite{Zhang2020DIALOGPTL} is a dialogue generation model via a language modeling target. SimCSE \cite{gao-etal-2021-simcse} uses Dropout to construct positive pairs and is further pre-trained on the same TOD corpus. TOD-BERT \cite{Wu2020TODBERTPN} uses a contrastive response selection objective by treating a response utterance and its dialogue context as positive pair. DSE \cite{Zhou2022LearningDR} takes consecutive utterances of the same dialogue as positive pairs.\footnote{We choose the unsupervised version of DSE in the original paper as our baseline for fair comparison.} Note that we focus on the unsupervised TOD pre-training, so we don't compare supervised methods using labeled NLI datasets \cite{Williams2018ABC} or dialogue act labels \cite{He2022SPACE2TS}.

\noindent\textbf{Pre-trainging Details} We train FutureTOD with a batch size of 32 and a maximum input length set of 512, respectively. Both the teacher and student models are initialized by BERT-base-uncased. Adam optimizer and a linear learning rate scheduler are employed for optimization with an initial learning rate of 5e-5 and a dropout ratio of 0.2. The mask ratio, teacher's update frequency, and the number of layers representations are set to 15\%, 10 epoch, and 12 respectively. Experiments take 3 days with an early-stopped strategy based on perplexity scores of a held-out development conducted on eight NVIDIA Tesla A100 GPUs. The average length of context and response are 86.04 and 48.10 tokens respectively. The average number of utterances in context and response are 5.95 and 3.48 respectively. We use the pre-trained BERT-MLM and pre-trained TOD-BERT released by the original paper \cite{Wu2020TODBERTPN}, and pre-trained DSE model released by \citet{Zhou2022LearningDR} respectively. We use Dropout to construct positive pairs to re-implement SimCSE \cite{gao-etal-2021-simcse}. For a fair comparison, we augment every single utterance obtained through Dropout on our pre-training corpora.

\subsection{Finetuning Setting}
We finetune these pre-trained LMs on the following four core downstream tasks in a task-oriented system: intent recognition, dialogue state tracking, dialogue act prediction, and response selection. Following \citet{Wu2020TODBERTPN}, we only use the LMs and avoid adding too many additional components except a classification head. We use the representation of the [CLS] token as the utterance representation here. Additionally, we provide the performance of the mean pooling in Appendix \ref{sec:mean_pool}. For fair comparison, we use the same architecture for all the baselines. Along with the full data setting, we also randomly sample a few labeled training examples as the few-shot learning settings. More hyperparameters details can be seen in Appendix \ref{sec:fintunedetails}.


\textbf{Intent Recognition} is a multi-class classification task, where 
the model predicts one intent label given an input sentence. We use the 
[CLS] embeddings as the dialogue representation and a softmax classification head. The model is trained with cross-entropy loss. We use OOS \cite{larson-etal-2019-evaluation} intent dataset, which covers 151 intent classes over ten domains, including 150 in-domain intents and one out-of-domain (OOD) intent. We treat the OOD intent as an additional class following TOD-BERT. We report classification accuracy and recall.

\begin{table}[t]
\centering
\resizebox{0.50\textwidth}{!}{
\begin{tabular}{c|l|l|l|l|l}
\hline
\multicolumn{1}{l|}{}                                                      & \multicolumn{1}{l|}{\textbf{Model}} & \multicolumn{1}{l|}{\begin{tabular}[c]{@{}l@{}}Acc\\ (all)\end{tabular}} & \multicolumn{1}{l|}{\begin{tabular}[c]{@{}l@{}}Acc\\ (in)\end{tabular}} & \multicolumn{1}{l|}{\begin{tabular}[c]{@{}l@{}}Acc\\ (out)\end{tabular}} & \multicolumn{1}{l}{\begin{tabular}[c]{@{}l@{}}Recall\\ (out)\end{tabular}} \\ \hline
\multirow{6}{*}{\textbf{1-Shot}}                                           & BERT                       & 29.3\%                                                                   & 35.7\%                                                                  & 81.3\%                                                                   & 0.4\%                                                                      \\
                                                                           & BERT-mlm                   & 38.9\%                                                                   & 47.4\%                                                                  & 81.6\%                                                                   & 0.5\%                                                                      \\
                                                                           & SimCSE                     & 29.9\%                                                                   & 36.4\%                                                                  & 81.7\%                                                         & 0.6\%                                                                      \\
                                                                           & TOD-BERT                   & 42.5\%                                                                   & 52.0\%                                                                  & 81.7\%                                                                   & 0.1\%                                                                      \\
                                                                           & DSE                        & 42.3\%                                                                   & 51.7\%                                                                  & 81.8\%                                                                   & 0.4\%                                                             \\
                                                                           & FutureTOD                  & \textbf{43.1\%}*                                                          & \textbf{52.2\%}                                                         & \textbf{81.8}\%                                                                   & \textbf{2.1\%}*                                                                     \\ \hline
\multirow{6}{*}{\textbf{10-Shot}}                                                   & BERT                       & 75.5\%                                                                   & 88.6\%                                                                  & 84.7\%                                                                   & 16.5\%                                                                     \\
                                                                           & BERT-mlm                   & 76.6\%                                                                   & 90.5\%                                                                  & 84.3\%                                                                   & 14.0\%                                                                     \\
                                                                           & SimCSE                     & 74.5\%                                                                   & 88.9\%                                                                  & 83.5\%                                                                   & 9.6\%                                                                      \\
                                                                           & TOD-BERT                   & 77.3\%                                                                   & \textbf{91.0\%}                                                         & 84.5\%                                                                   & 15.3\%                                                                     \\
                                                                           & DSE                        & 77.8\%                                                                   & 90.8\%                                                                  & 85.2\%                                                                   & 19.1\%                                                                     \\
                                                                           & FutureTOD                  & \textbf{78.1\%}                                                          & 90.8\%                                                                  & \textbf{85.5\%}*                                                          & \textbf{20.5\%}*                                                            \\ \hline
\multirow{6}{*}{\begin{tabular}[c]{@{}c@{}}\textbf{Full}\\ (\textbf{100-shot})\end{tabular}} & BERT                       & 84.9\%                                                                   & 95.8\%                                                                  & 88.1\%                                                                   & 35.6\%                                                                     \\
& DialoGPT                   & 83.9\%                                                                   & 95.5\%                                                                  & 87.6\%                                                                   & 32.1\%                                                                     \\
                                                                           & BERT-mlm                   & 85.9\%                                                                   & 96.1\%                                                                  & 89.5\%                                                                   & 46.3\%                                                                     \\
                                                                           & SimCSE                     & 82.3\%                                                                   & 94.7\%                                                                  & 86.6\%                                                                   & 26.6\%                                                                     \\
                                                                           & TOD-BERT                   & 86.6\%                                                                   & \textbf{96.2\%}                                                         & 89.9\%                                                                   & 43.6\%                                                                     \\
                                                                           & DSE                        & 84.3\%                                                                   & 95.8\%                                                                  & 87.7\%                                                                   & 32.5\%                                                                     \\
                                                                           & FutureTOD                  & \textbf{87.2\%}*                                                          & 96.0\%                                                                  & \textbf{90.0\%}                                                          & \textbf{47.6\%}*                                                            \\ \hline
\end{tabular}
}

\caption{Intent recognition results on the OOS dataset. Acc(all), Acc(in), Acc(out) denotes the overall accuracy, in-domain intent accuracy and out-of-domain intent accuracy. The numbers with * are significant using t-test with $p < 0.01$.}
\label{main_intent}

\end{table}

\textbf{Dialogue State Tracking} is regarded as a multi-class classification task based on a pre-defined ontology. We use dialogue history as input and predict slot values for each (domain, slot) pair at each dialogue turn. The model is trained with cross-entropy loss summed over all the pairs. We use a widely-used TOD dataset MWOZ 2.1 \cite{Budzianowski2018MultiWOZA} across seven different domains. We report joint goal accuracy and slot accuracy. The former considers true if and only if all the predicted values exactly match its ground truth values at each dialogue turn while the latter individually compares each (domain, slot, value) triplet to its ground truth label. Joint goal accuracy is the main metric.

\textbf{Dialogue Act Prediction} is a multi-label classification task where the model takes dialogue history as input and predicts the system actions. The model is trained with binary cross-entropy loss summed over all the actions. For prediction, we set the threshold to 0.5. We use two datasets MWOZ \cite{Budzianowski2018MultiWOZA} and DSTC2 \cite{Henderson2014TheSD}. Following \citet{Wu2020TODBERTPN}, we use the same data preprocessing to uniform the original dialogue acts to a general format. We report micro-F1 and macro-F1 scores for the dialogue act prediction task.

\begin{table*}[t]
\centering
\resizebox{1.0\textwidth}{!}{
\begin{tabular}{l|ll|ll|ll|ll|ll}
\hline
\multirow{2}{*}{Model}          & \multicolumn{2}{c|}{1\% Data}                               & \multicolumn{2}{c|}{5\% Data}          & \multicolumn{2}{c|}{10\% Data}                                                 & \multicolumn{2}{c|}{25\% Data}                                                 & \multicolumn{2}{c}{Full Data}                                               \\
                                & \multicolumn{1}{c}{\textbf{Joint Acc}} & \textbf{Slot Acc} & \textbf{Joint Acc} & \textbf{Slot Acc} & \multicolumn{1}{l}{\textbf{Joint Acc}} & \multicolumn{1}{l}{\textbf{Slot Acc}} & \multicolumn{1}{|l}{\textbf{Joint Acc}} & \multicolumn{1}{l}{\textbf{Slot Acc}} & \multicolumn{1}{|l}{\textbf{Joint Acc}} & \multicolumn{1}{l}{\textbf{Slot Acc}} \\ \hline
\multicolumn{1}{l|}{BERT}      & \multicolumn{1}{c}{6.4\%}              & 84.4\%            & 19.6\%             & 92.0\%            & 32.9\%                                 & 94.7\%                                & 40.8\%                                 & 95.8\%                                & 45.6\%                                 & 96.6\%                                \\
\multicolumn{1}{l|}{BERT-mlm}  & \multicolumn{1}{c}{9.9\%}              & \textbf{86.6}\%   & 28.1\%             & 93.9\%            & 39.5\%                                 & 95.6\%                                & 44.0\%                                 & 96.4\%                                & 47.7\%                                 & 96.8\%                                \\
\multicolumn{1}{l|}{SimCSE}    & \multicolumn{1}{c}{7.4\%}              & 84.8\%            & 21.1\%             & 91.6\%            & 35.6\%                                 & 95.0\%                                & 43.8\%                                 & 96.3\%                                & 48.0\%                                 & 96.8\%                                \\
\multicolumn{1}{l|}{TOD-BERT}  & \multicolumn{1}{c}{8.0\%}              & 85.3\%            & 28.6\%             & 93.8\%            & 37.0\%                                 & 95.2\%                                & 44.3\%                                 & 96.3\%                                & 48.0\%                                 & 96.9\%                                \\
\multicolumn{1}{l|}{DSE}       & \multicolumn{1}{c}{9.8\%}              & 86.3\%            & 23.8\%             & 93.0\%            & 37.8\%                                 & 95.5\%                                & 43.4\%                                 & 96.3\%                                & 49.9\%                                 & 97.0\%                                \\
\multicolumn{1}{l|}{FutureTOD} & \multicolumn{1}{c}{\textbf{9.9}\%}     & 85.5\%            & \textbf{29.1}\%*   & \textbf{94.1}\%*  & \textbf{40.7}\%*                       & \textbf{95.8}\%                       & \textbf{45.7}\%*                       & \textbf{96.5}\%                       & \textbf{50.4}\%*                       & \textbf{97.1}\%                 \\\hline     
\end{tabular}
}

\caption{Dialogue state tracking results on MWOZ 2.1. We report joint goal accuracy (Joint Acc) and slot accuracy (Slot Acc) for the full data and few-shot settings. The numbers with * are significant using t-test with $p < 0.01$.}
\label{main_dst}

\end{table*}

\begin{table}[t]
\centering
\resizebox{0.50\textwidth}{!}{
\begin{tabular}{c|l|ll|ll}
\hline
\multirow{2}{*}{}                   & \multicolumn{1}{c|}{\multirow{2}{*}{\textbf{Model}}} & \multicolumn{2}{c|}{\textbf{MWOZ}}                 & \multicolumn{2}{c}{\textbf{DSTC2}}                  \\ \cline{3-6} 
                                    & \multicolumn{1}{c|}{}                                & \multicolumn{1}{c|}{micro-F1}        & macro-F1        & \multicolumn{1}{c|}{micro-F1}        & macro-F1        \\ \hline
\multirow{6}{*}{\textbf{1\% Data}}  & BERT                                                 & \multicolumn{1}{l|}{84.0\%}          & 66.7\%          & \multicolumn{1}{l|}{77.1\%}          & 25.8\%          \\
                                    & BERT-mlm                                             & \multicolumn{1}{l|}{87.5\%}          & 73.3\%          & \multicolumn{1}{l|}{79.6\%}          & 26.4\%          \\
                                    & SimCSE                                               & \multicolumn{1}{l|}{81.0\%}          & 62.1\%          & \multicolumn{1}{l|}{78.9\%}          & 27.3\%          \\

                                    & TOD-BERT                                         & \multicolumn{1}{l|}{86.9\%}          & 72.4\%          & \multicolumn{1}{l|}{82.9\%}          & 28.0\%          \\
                                    & DSE                                                  & \multicolumn{1}{l|}{82.9\%}          & 65.1\%          & \multicolumn{1}{l|}{72.4\%}          & 21.4\%          \\
                                    & FutureTOD                                            & \multicolumn{1}{l|}{\textbf{87.9\%}*} & \textbf{75.0\%}* & \multicolumn{1}{l|}{\textbf{83.7\%}*} & \textbf{31.0\%}* \\ \hline
\multirow{6}{*}{\textbf{10\% Data}} & BERT                                                 & \multicolumn{1}{l|}{89.7\%}          & 78.4\%          & \multicolumn{1}{l|}{88.2\%}          & 34.8\%          \\
                                    & BERT-mlm                                             & \multicolumn{1}{l|}{90.1\%}          & 78.9\%          & \multicolumn{1}{l|}{91.8\%}          & 39.4\%          \\
                                    & SimCSE                                               & \multicolumn{1}{l|}{89.6\%}          & 77.8\%          & \multicolumn{1}{l|}{92.3\%}          & 40.5\%          \\
                                    & TOD-BERT                                         & \multicolumn{1}{l|}{90.2\%}          & 79.6\%          & \multicolumn{1}{l|}{90.6\%}          & 38.8\%          \\
                                    & DSE                                                  & \multicolumn{1}{l|}{89.9\%}          & 79.4\%          & \multicolumn{1}{l|}{91.1\%}          & 39.0\%          \\
                                    & FutureTOD                                            & \multicolumn{1}{l|}{\textbf{91.0\%}*} & \textbf{80.5\%}* & \multicolumn{1}{l|}{\textbf{93.6\%}*} & \textbf{40.9\%} \\ \hline
\multirow{7}{*}{\textbf{Full Data}} 
                                    & BERT                                                 & \multicolumn{1}{l|}{91.4\%}          & 79.7\%          & \multicolumn{1}{l|}{92.3\%}          & 40.1\%          \\

& DialoGPT                                             & \multicolumn{1}{l|}{91.2\%}          & 79.7\%          & \multicolumn{1}{l|}{93.8\%}          & 42.1\%          \\
                                    & BERT-mlm                                             & \multicolumn{1}{l|}{91.7\%}          & 79.9\%          & \multicolumn{1}{l|}{90.9\%}          & 39.9\%          \\
                                    & SimCSE                                               & \multicolumn{1}{l|}{91.6\%}          & 80.3\%          & \multicolumn{1}{l|}{91.5\%}          & 39.6\%          \\

                                    & TOD-BERT                                         & \multicolumn{1}{l|}{91.7\%}          & 80.6\%          & \multicolumn{1}{l|}{93.8\%}          & 41.3\%          \\
                                    & DSE                                                  & \multicolumn{1}{l|}{91.7\%}          & 81.3\%          & \multicolumn{1}{l|}{92.6\%}          & 40.2\%          \\
                                    & FutureTOD                                            & \multicolumn{1}{l|}{\textbf{92.0\%}} & \textbf{81.9\%}* & \multicolumn{1}{l|}{\textbf{94.6\%}*} & \textbf{44.6\%}* \\ \hline
\end{tabular}
}
\caption{Dialogue act prediction results on MWOZ and DSTC2. The numbers with * are significant using t-test with $p < 0.01$.}
\label{main_act}
\end{table}

\begin{table}[t]
\centering
\resizebox{0.50\textwidth}{!}{
\begin{tabular}{c|l|ll|ll}
\hline
                                     & \multicolumn{1}{c|}{}                                 & \multicolumn{2}{c|}{\textbf{MWOZ}}                                                                 & \multicolumn{2}{c}{\textbf{DSTC2}}                                                \\ \cline{3-6} 
\multirow{-2}{*}{}                   & \multicolumn{1}{c|}{\multirow{-2}{*}{\textbf{Model}}} & \multicolumn{1}{c|}{1-to-100}                                & 3-to-100                                & \multicolumn{1}{c|}{1-to-100}                       & 3-to-100                       \\ \hline
                                     & BERT                                                  & \multicolumn{1}{l|}{7.8\%}                                   & 20.5\%                                  & \multicolumn{1}{l|}{3.7\%}                          & 9.6\%                          \\
                                     & BERT-mlm                                              & \multicolumn{1}{l|}{13.0\%}                                  & 34.6\%                                  & \multicolumn{1}{l|}{12.5\%}                         & 24.9\%                         \\
                                     & SimCSE                                                & \multicolumn{1}{l|}{17.2\%}                                  & 32.6\%                                  & \multicolumn{1}{l|}{27.6\%}                         & 46.4\%                         \\
                                     & TOD-BERT                                          & \multicolumn{1}{l|}{-}               & -               & \multicolumn{1}{l|}{37.5\%} & 55.9\% \\
                                     & DSE                                                   & \multicolumn{1}{l|}{7.9\%}                                   & 21.2\%                                  & \multicolumn{1}{l|}{2.4\%}                          & 6.1\%                          \\
\multirow{-6}{*}{\textbf{1\% Data}}  & FutureTOD                                             & \multicolumn{1}{l|}{\textbf{35.8\%}*} & \textbf{53.5\%}* & \multicolumn{1}{l|}{\textbf{39.5\%}*}                & \textbf{64.0\%}*                \\ \hline
                                     & BERT                                                  & \multicolumn{1}{l|}{20.9\%}                                  & 45.4\%                                  & \multicolumn{1}{l|}{8.9\%}                          & 21.4\%                         \\
                                                                          & BERT-mlm                                              & \multicolumn{1}{l|}{22.3\%}          & 48.7\%          & \multicolumn{1}{l|}{19.0\%} & 33.8\% \\
                                     & SimCSE                                                & \multicolumn{1}{l|}{37.2\%}          & 60.6\%          & \multicolumn{1}{l|}{42.0\%} & 63.5\% \\

                                     & TOD-BERT                                          & \multicolumn{1}{l|}{-}               & -               & \multicolumn{1}{l|}{49.7\%}                         & 66.6\%                         \\
                                     & DSE                                                   & \multicolumn{1}{l|}{24.8\%}          & 49.4\%          & \multicolumn{1}{l|}{42.0\%} & 59.7\% \\
\multirow{-6}{*}{\textbf{10\% Data}} & FutureTOD                                             & \multicolumn{1}{l|}{\textbf{50.0\%}*}                         & \textbf{72.8\%}*                         & \multicolumn{1}{l|}{\textbf{51.3\%}*}                & \textbf{70.0\%}*                \\ \hline
                                     & BERT                                                  & \multicolumn{1}{l|}{47.5\%}          & 75.5\%          & \multicolumn{1}{l|}{46.6\%} & 62.1\% \\
& DialoGPT                                              & \multicolumn{1}{l|}{35.7\%}                                  & 64.1\%                                  & \multicolumn{1}{l|}{39.8\%}                         & 57.1\%                         \\
                                     & BERT-mlm                                              & \multicolumn{1}{l|}{48.1\%}                                  & 74.3\%                                  & \multicolumn{1}{l|}{50.0\%}                         & 65.1\%                         \\
                                     & SimCSE                                                & \multicolumn{1}{l|}{64.2\%}                                  & 85.4\%                                  & \multicolumn{1}{l|}{55.6\%}                         & 70.5\%                         \\
                                     & TOD-BERT                                          & \multicolumn{1}{l|}{65.8\%}          & 87.0\%          & \multicolumn{1}{l|}{56.8\%} & 70.6\% \\
                                     & DSE                                                   & \multicolumn{1}{l|}{63.3\%}                                  & 85.3\%                                  & \multicolumn{1}{l|}{58.3\%}                         & 72.0\%                         \\
\multirow{-7}{*}{\textbf{Full Data}} & FutureTOD                                             & \multicolumn{1}{l|}{\textbf{68.5\%}*}                         & \textbf{87.9\%}*                         & \multicolumn{1}{l|}{\textbf{58.4\%}}                & \textbf{72.6\%}*                \\ \hline
\end{tabular}
}

\caption{Response selection evaluation results on MWOZ and DSTC2 for 1\%, 10\% and full data setting. We report 1-to-100 and 3-to-100 accuracy, which represents the ratio of the ground-truth response being ranked at the top-1 or top-3 given 100 candidates. The numbers with * are significant using t-test with $p < 0.01$.}
\label{main_rs}

\end{table}

\textbf{Response Selection} is a ranking task where the model selects the most relevant response from a candidate pool given an input dialogue history. We use a shared pre-trained LM to encode the dialogue and each response respectively and compute its cosine similarity score. We randomly sample several system responses from the corpus as negative samples. In our experiments, we set batch size equals to 25 for all the models. We also use MWOZ and DSTC2 as our evaluation datasets. We use k-to-100 accuracy as the metric. For each history, we combine its ground-truth response with 99 randomly sampled responses and rank these 100 responses based on their similarities with the query in the embedding space. The k-to-100 accuracy represents the ratio of the ground-truth response being ranked at the top-k.

\subsection{Main Results}

\textbf{Intent Recognition}
We evaluate our FutureTOD on the intent recognition dataset OOS, including in-domain (IND) and out-of-domain (OOD) in Table \ref{main_intent}. We find FutureTOD outperforms all the baselines on 10 of 12 metrics, especially with significant improvements in overall accuracy and OOD metrics. SimCSE (82.3\% Acc(all)) is even worse than the original BERT (84.9\% Acc(all)) in the full setting. Moreover, the 1.5 drop of Acc(out) is more significant than 1.1 of Acc(in), demonstrating that SimCSE ignores intrinsic dialogue structures and fails to model the relations between each utterance in the same dialogue. We also find TOD-BERT achieves comparable performance on Acc(in) except Recall(out), indicating the robustness of our method. Surprisingly, a recent strong baseline DSE performs poorly in the full setting. We argue the assumption that consecutive utterances represent similar semantics may fail in practical dialogues. Generally, FutureTOD achieves comparable or higher performance on in-domain intent accuracy, but significant improvements on out-of-domain accuracy, which proves the robustness and generalization ability of our method.

\textbf{Dialogue State Tracking}
Table \ref{main_dst} displays the results of dialogue state tracking on MWOZ 2.1. Our FutureTOD achieves state-of-the-art results on 9 of 10 metrics. We find our method obtains significant improvements on Joint Acc than Slot Acc, showing the superiority of modeling overall dialogue context. Although these baselines achieve fair results on each (domain, slot, value) triplet, we observe they tend to overfit to the easy slot value pairs with high accuracy but fail to recognize hard ones, leading to poor overall joint goal accuracy. For example, FutureTOD outperforms DSE by 0.1\% on Slot Acc but 0.5\% on Joint Acc.  All the results show the effectiveness of our method.

\begin{figure}[t]
    \centering
    \begin{adjustbox}{minipage=\linewidth,scale=1.0}
    \subfigure[MWOZ]{
        \includegraphics[width=0.48\textwidth]{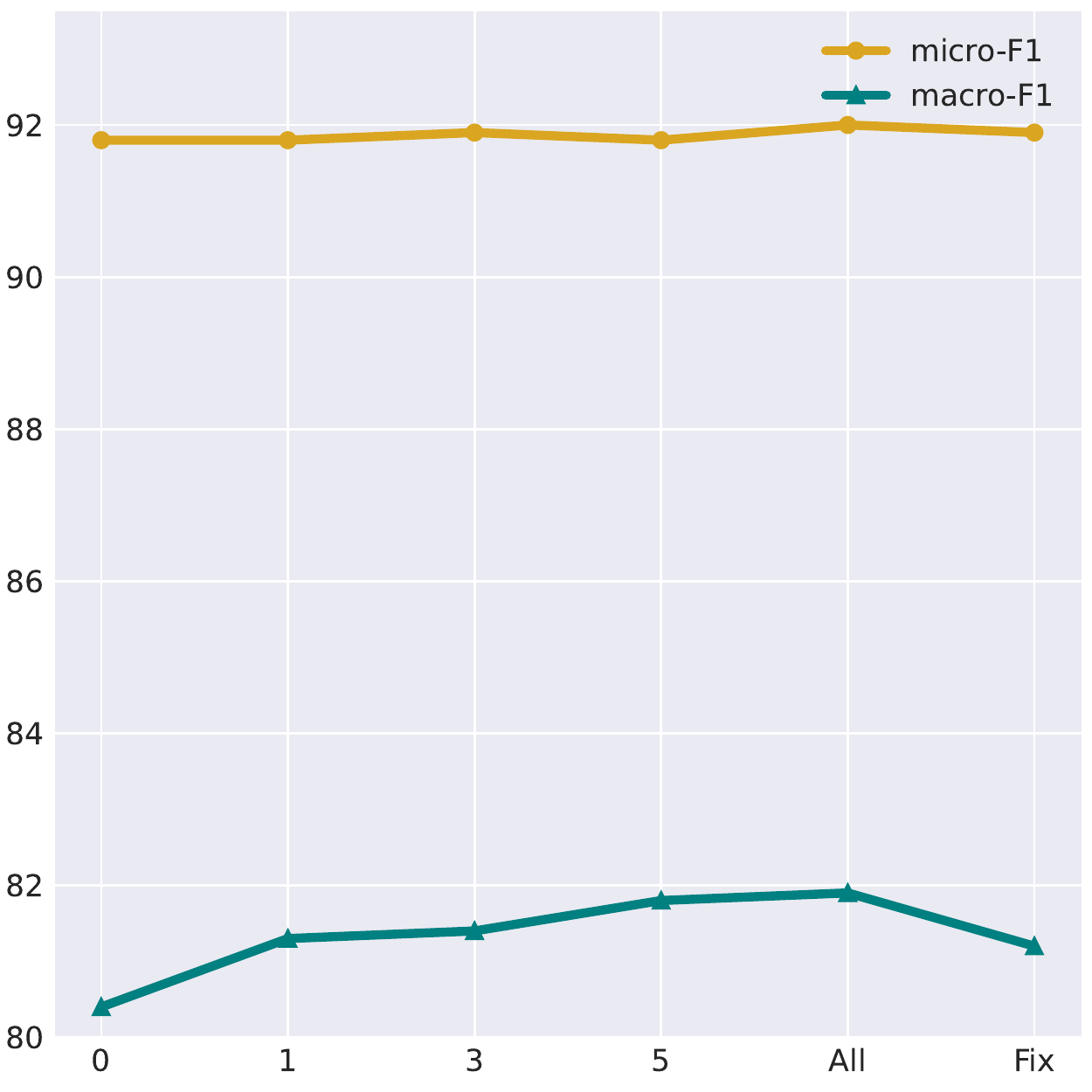}
    }
    \hspace{-0.4cm}
    \subfigure[DSTC2]{
        \includegraphics[width=0.48\textwidth]{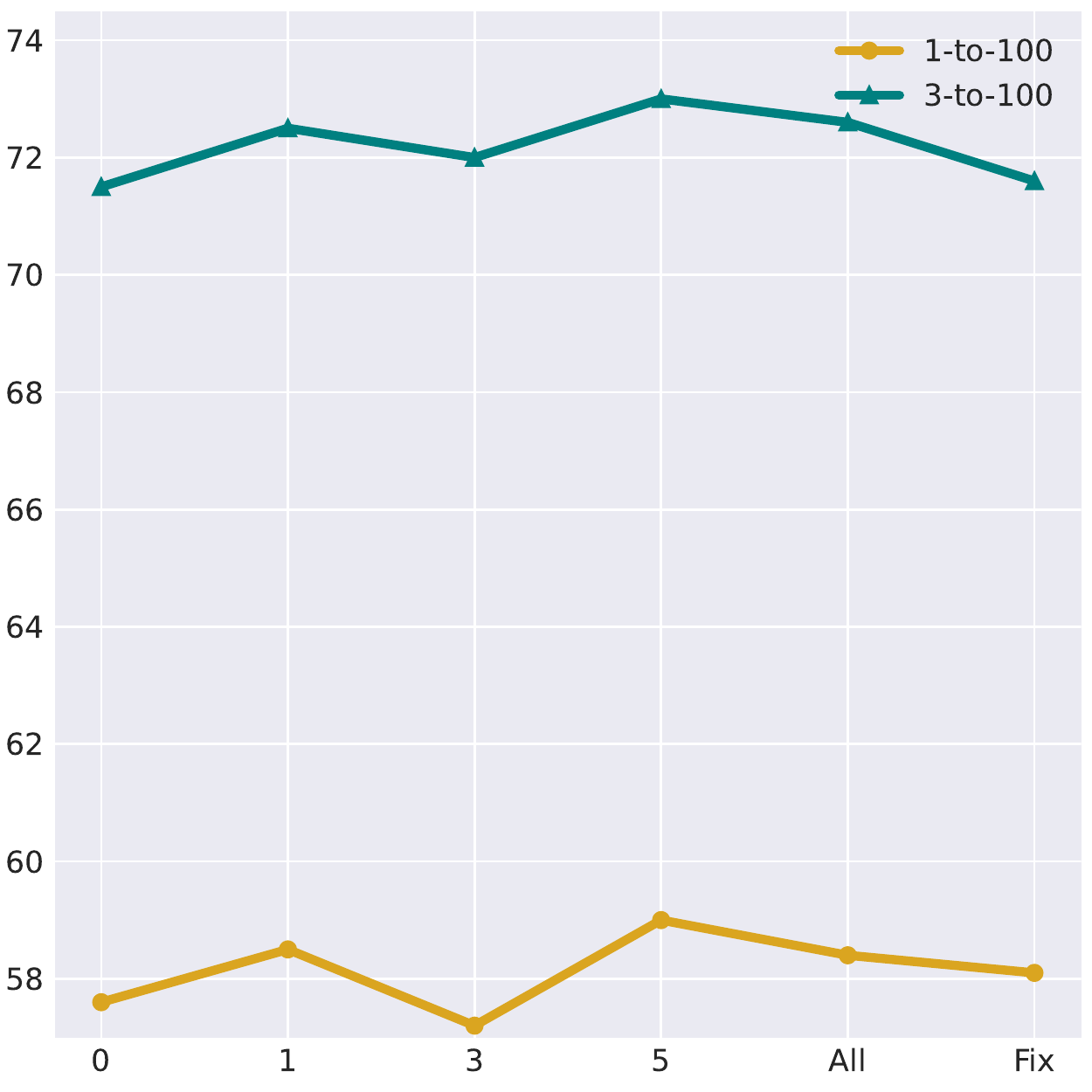}
    }

    \caption{Ablation study of max future lengths. We report the results of dialogue act prediction on MWOZ and response selection on DSTC2. The X-asix and Y-asix denotes the max future length and performance.}
    \label{fig:fut_len}
    \end{adjustbox}

\end{figure}

\textbf{Dialogue Act Prediction}
Table \ref{main_act} shows the results of dialogue act prediction on MWOZ and DSTC2. Our FutureTOD achieves state-of-the-art results on all the metrics. We find our method obtains comparable performance only using 10\% data than the baselines using 100\% data, which verifies the superior few-shot learning capability. We find DSE performs poorly in the 1\% data setting because the original DSE uses one utterance as the query and lacks the ability of modeling long context. In contrast, our model achieves consistent performance in all the settings, showing better generalization ability than previous baselines.

\textbf{Response Selection}
Table \ref{main_rs} displays the results of response selection on MWOZ and DSTC2. Our FutureTOD achieves state-of-the-art results on all the metrics. Besides, we find the improvements in the 1\% data setting are more significant than the full data. Note that TOD-BERT uses the response contrastive learning as the pre-training objective on full MWOZ training data so we don't report its results of few-shot learning. However, our method still significantly outperforms TOD-BERT on DSTC2 without using response selection loss. It proves FutureTOD learns generalized dialogue representations by distilling future knowledge to pre-trained models and performs well on downstream tasks.

Overall, FutureTOD achieves state-of-the-art results for most of the downstream tasks while existing dialogue pre-trained models fail in specific tasks. The results demonstrate our pre-training method has strong generalization capability for diverse dialogue tasks. The results on out-of-domain intent recognization also prove its robustness.

\section{Qualitative Analysis}
\subsection{Hyper-parameter Analysis}

\textbf{Effect of Max Future Length} We randomly select a part of future utterances ranging from 1 to the max future length $P$. To explore the effect of different max future lengths, we set the $P$ to 1, 3, 5, and $All$ respectively. \footnote{If the real length of total future utterances is lower than the given max limit, we just randomly select from the whole future.} If the $P=All$, we can randomly select any length of utterances from the whole future utterances. For comparison, we also add a baseline $P=Fix$ which must use the whole future utterances together. For example, if we have 5 future utterances $F=\{S_{t}, U_{t+1}, S_{t+1}, U_{t+2}, S_{t+2}\}$. When $P=3$, we can select any length no longer than 3, such as $\{S_{t}\}$ or $\{S_{t}, U_{t+1}, S_{t+1}\}$; When $P = All$, we can select any length of future from the 5 utterances, that is $\{S_{t}\}$ or $\{S_{t}, U_{t+1}, S_{t+1}\}$ or $F$; When $P=Fix$, we can only select $F$. Figure \ref{fig:fut_len} shows that FutureTOD generally gets improvements with increasing $P$. We argue that more future turns make the model learn comprehensive knowledge. We also observe that directly using all the future utterances like $P=Fix$ can't bring further improvements because diverse future knowledge with different granularity also makes an effect. An intuitive explanation is that too long future utterances possibly cause bias to a short dialogue context. Assuming a context only contains a single utterance but we always use ten, even more, future utterances to distill knowledge, the representation of the context will overfit to the future. Randomly selecting future information plays a role similar to Dropout \cite{Srivastava2014DropoutAS}. We leave more complicated selection strategies to future work, such as adaptively selecting the future for different lengths of context. We also conducted experiments using a teacher model that only encodes the future. However, the model's performance is poor. For detailed analysis, please refer to the Appendix \ref{sec:only_the_future}

\begin{figure}[t]
    \centering
    \begin{adjustbox}{minipage=\linewidth,scale=1.0}
    \subfigure[MWOZ]{
        \includegraphics[width=0.48\textwidth]{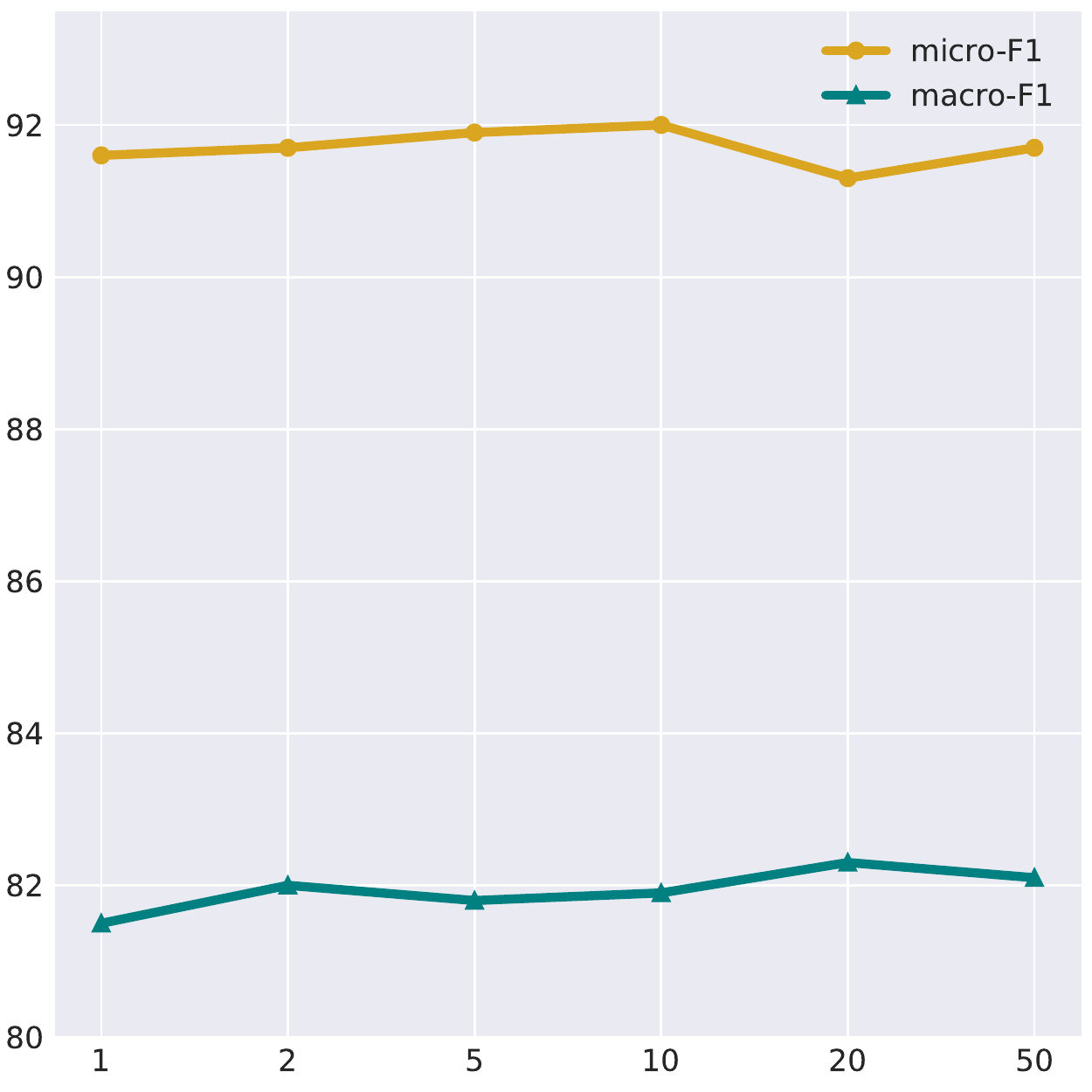}
    }
    \hspace{-0.5cm}
    \subfigure[DSTC2]{
        \includegraphics[width=0.48\textwidth]{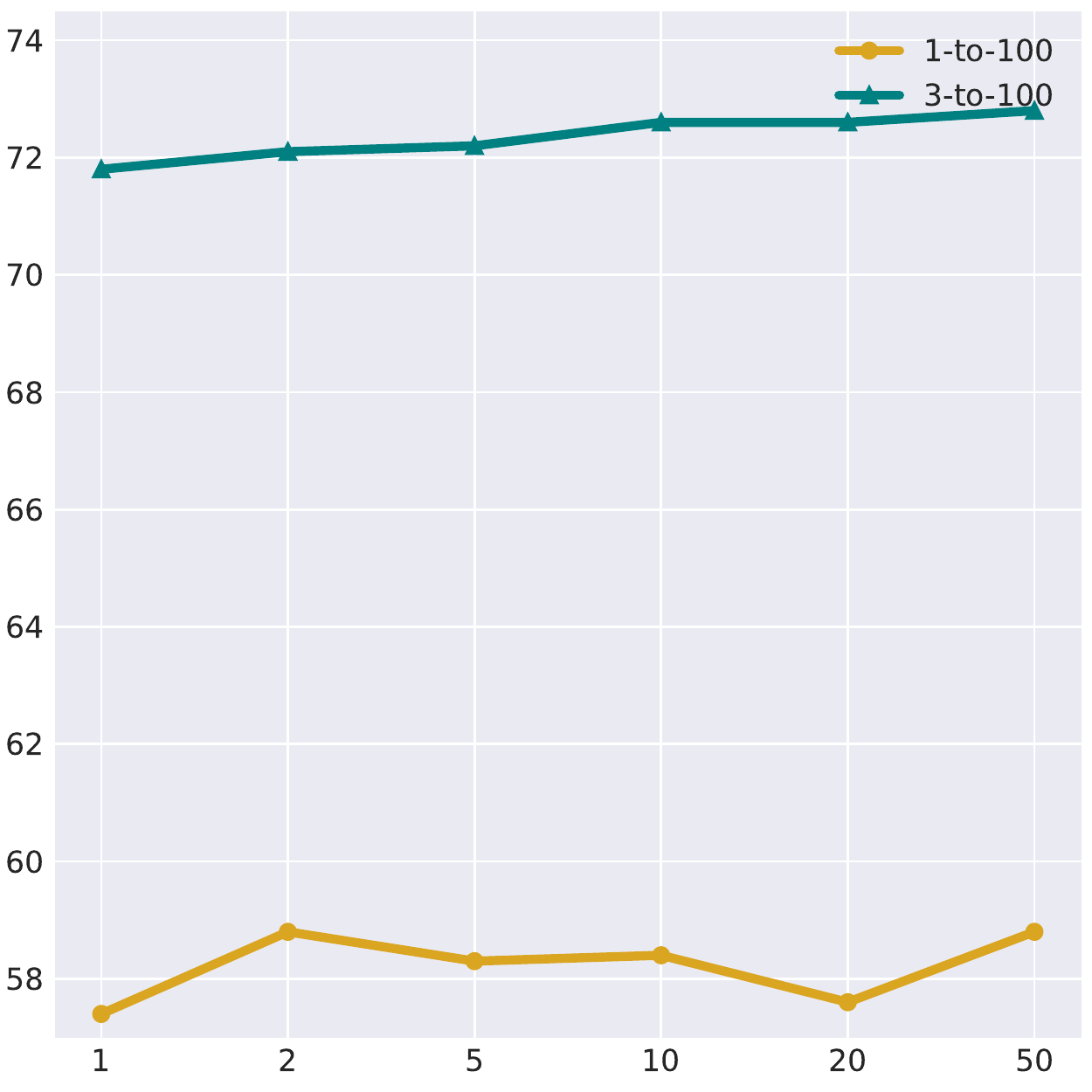}
    }

    \caption{Ablation study of the teacher's update frequency. We conduct dialogue act prediction on MWOZ and response selection on DSTC2. The X-asix and Y-asix denotes update frequency and performance.}
    \label{fig:up_freq}
    \end{adjustbox}

\end{figure}

\begin{figure*}[t]
    \centering
    \subfigure[TOD-BERT]{
        \includegraphics[width=0.31\textwidth]{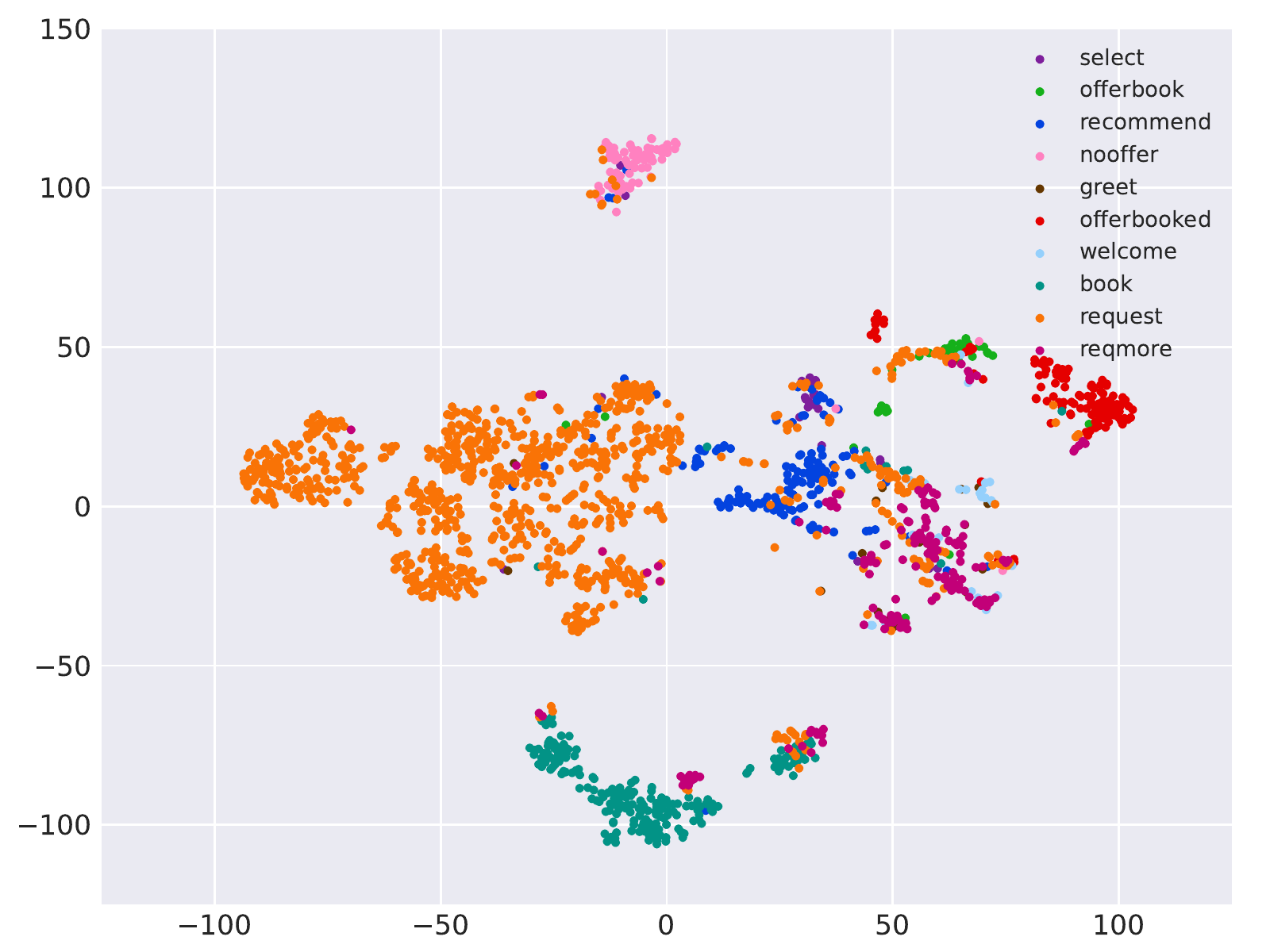}
    }
    \subfigure[DSE]{
        \includegraphics[width=0.31\textwidth]{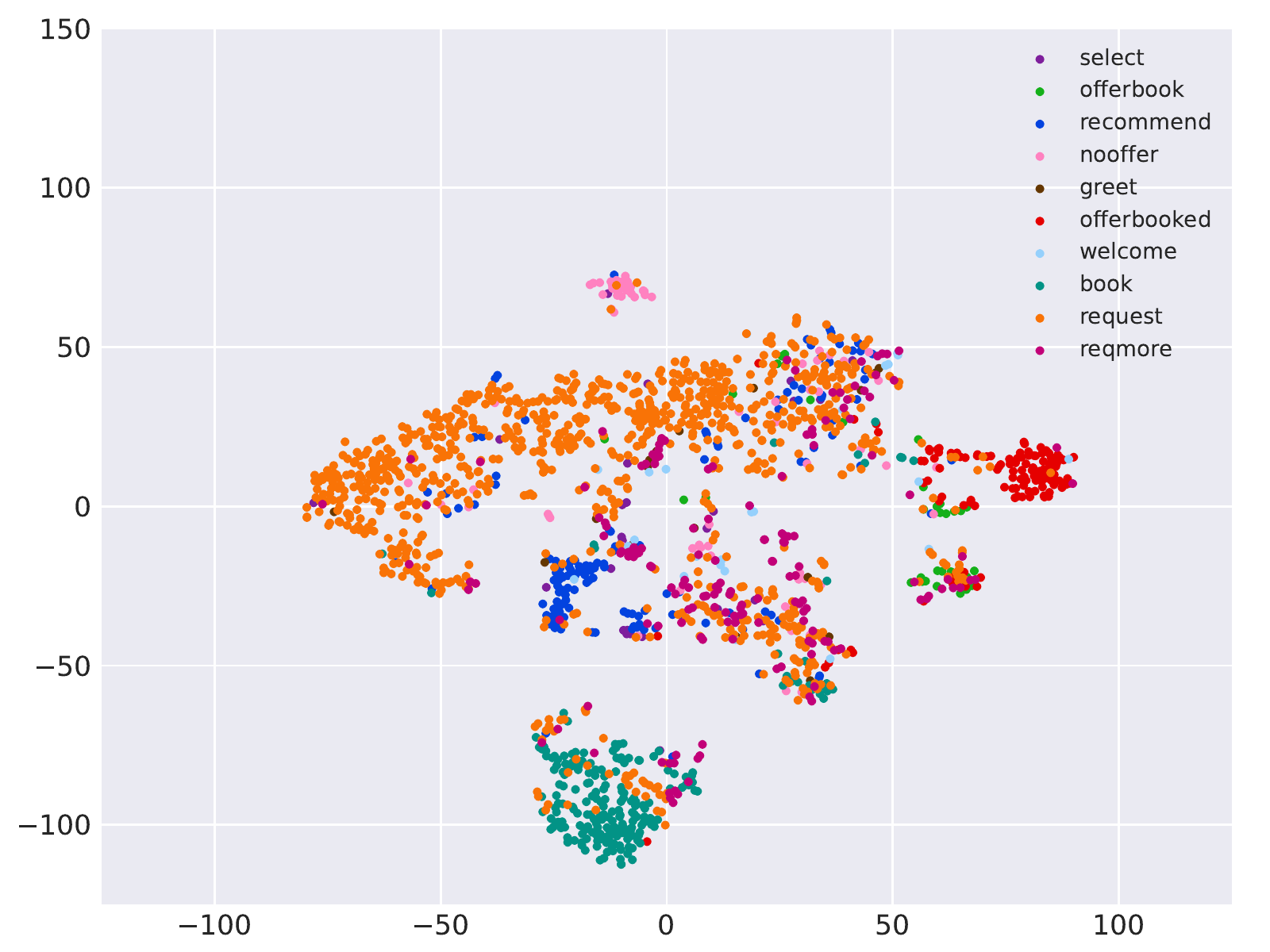}
    }
    \subfigure[FutureTOD]{
        \includegraphics[width=0.31\textwidth]{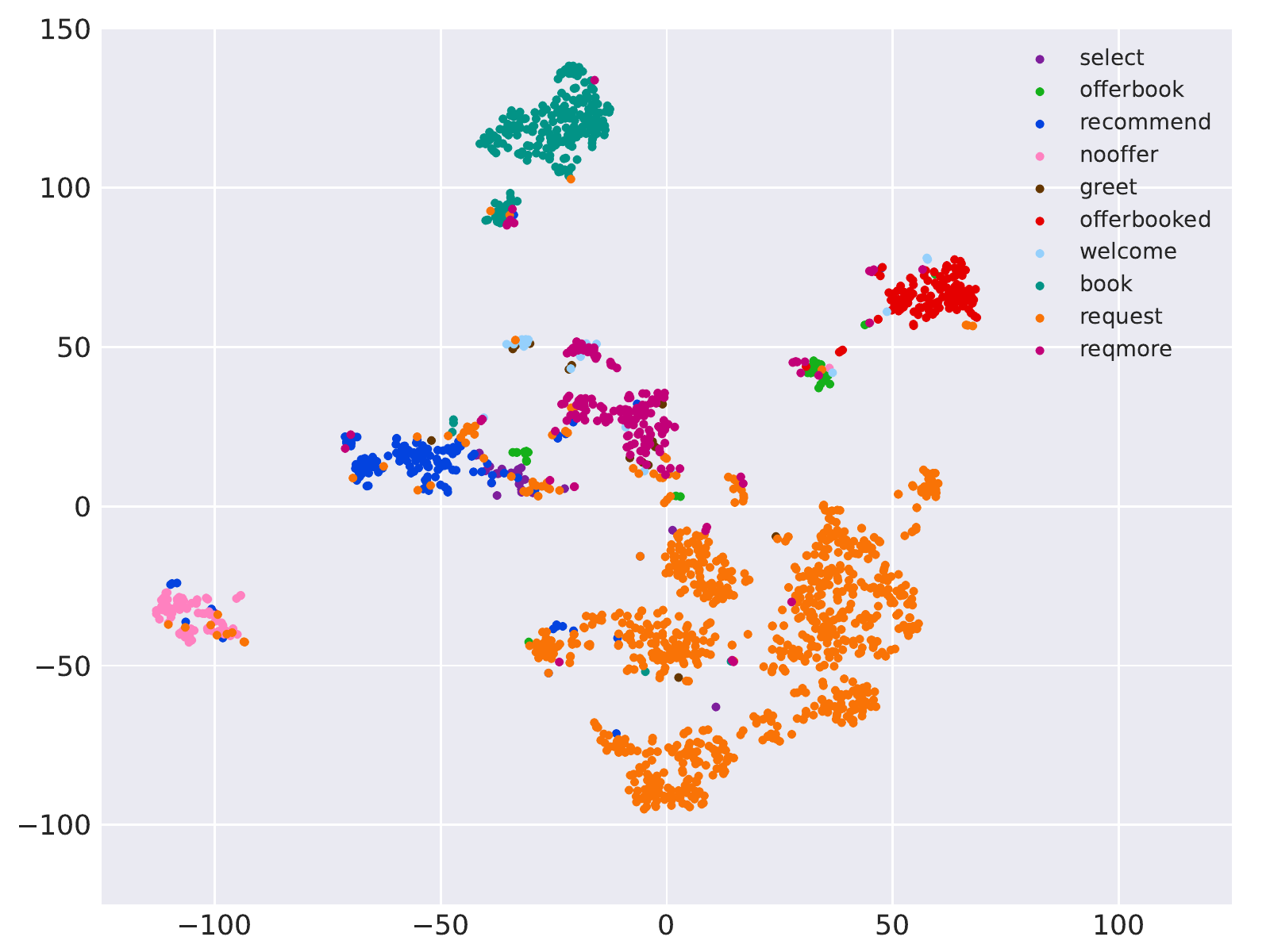}
    }

    \caption{ The  tSNE  visualization  of TOD-BERT, DSE and FutureTOD representations of system responses in the MWOZ test set. Different colors  represent different dialogue acts.}
    \label{fig:pca}

\end{figure*}

\label{e}
\textbf{Effect of Frequency of Updating Teacher} FutureTOD updates the teacher model using the student parameters every $E$ epoch. Figure \ref{fig:up_freq} shows the effect of updating frequency $E$. We find $E=10$ gets decent performance in general. We assume too small $E$ makes the teacher tightly close to the student and prone to collapse while too large $E$ can't produce a high-quality teacher model as learning signals and make the training slow. We also try other updating strategies such as momenta updating \cite{He2020MomentumCF} and non-constant $E$ but don't observe improvements. The simple strategy of updating every $E$ epoch is simple and robust.

\label{layer}
\textbf{Effect of Distillation Layers} We use the different top layers for the distillation loss Eq \ref{eq_dis} in Table \ref{tab:layer}. We find adding more layers for distilling future knowledge can significantly improve performance. It indicates that different types of features extracted at different layers enhance learning different granularity of future information and improve downstream tasks.

\begin{table}[t]
\centering
\resizebox{0.48\textwidth}{!}{
\begin{tabular}{c|cc|cc}
\hline
\multicolumn{1}{c|}{}                                 & \multicolumn{2}{c|}{\textbf{MWOZ}}                                                                                               & \multicolumn{2}{c}{\textbf{DSTC2}}                                                  \\ \cline{2-5} 
\multicolumn{1}{c|}{\multirow{-2}{*}{\textbf{Top-K Layer}}} & \multicolumn{1}{c|}{micro-F1}                                               & macro-F1                                               & \multicolumn{1}{c|}{1-to-100}                        & 3-to-100                        \\ \hline
1                                                & \multicolumn{1}{c|}{91.63\%}                                                & { 80.46\%}                         & \multicolumn{1}{c|}{58.08\%}                         & 72.11\%                         \\
3                                               & \multicolumn{1}{c|}{{ 91.60\%}}                         & { 80.49\%}                         & \multicolumn{1}{c|}{58.40\%}                         & 72.16\%                         \\
6                                                & \multicolumn{1}{c|}{91.75\%}                                                & { 81.02\%}                         & \multicolumn{1}{c|}{58.20\%}                         & \textbf{72.80\%}                         \\
9                                                & \multicolumn{1}{c|}{{ 91.72\%}} & { 80.89\%} & \multicolumn{1}{c|}{\textbf{58.51\%}} & 72.79\% \\
12                                              & \multicolumn{1}{c|}{\textbf{91.95\%}}                                                & { \textbf{81.92\%}}                         & \multicolumn{1}{c|}{58.41\%}                         & 72.60\%    
\\ \hline
\end{tabular}
}

\caption{Ablation study of using top-K layer representations for distillation. For example, $K=3$ denotes we use the top 3 layers of BERT-base to compute Eq \ref{dis}.}

\label{tab:layer}
\end{table}

\subsection{Visualization}

Figure \ref{fig:pca} shows the visualization of the system response representations of TOD-BERT, DSE and FutureTOD given the same input from the MWOZ test set. We use a pre-trained model to get [CLS] features and perform dimension reduction using the t-distributed stochastic neighbor embedding (tSNE). Different colors represent different dialogue act labels of the responses. We observe that FutureTOD builds compact and clearly separable dialogue representations for different clusters, which help distinguish semantically similar dialogues.


\subsection{Understanding Future Knowledge}

\begin{figure}[t]
    \centering
    \begin{adjustbox}{minipage=\linewidth,scale=1.0}
    \subfigure[MWOZ]{
        \includegraphics[width=0.48\textwidth]{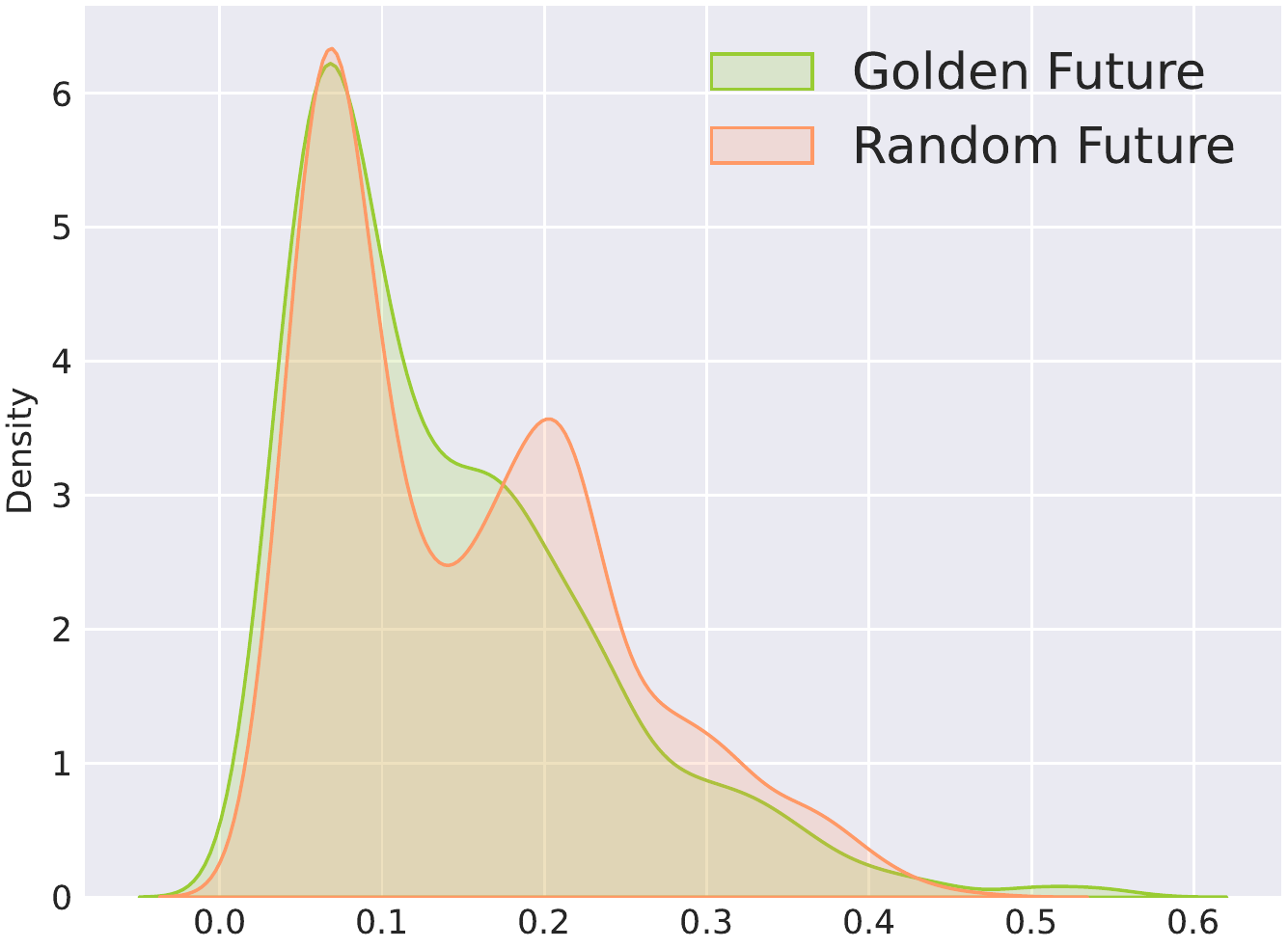}
    }
    \hspace{-0.5cm}
    \subfigure[DSTC2]{
        \includegraphics[width=0.48\textwidth]{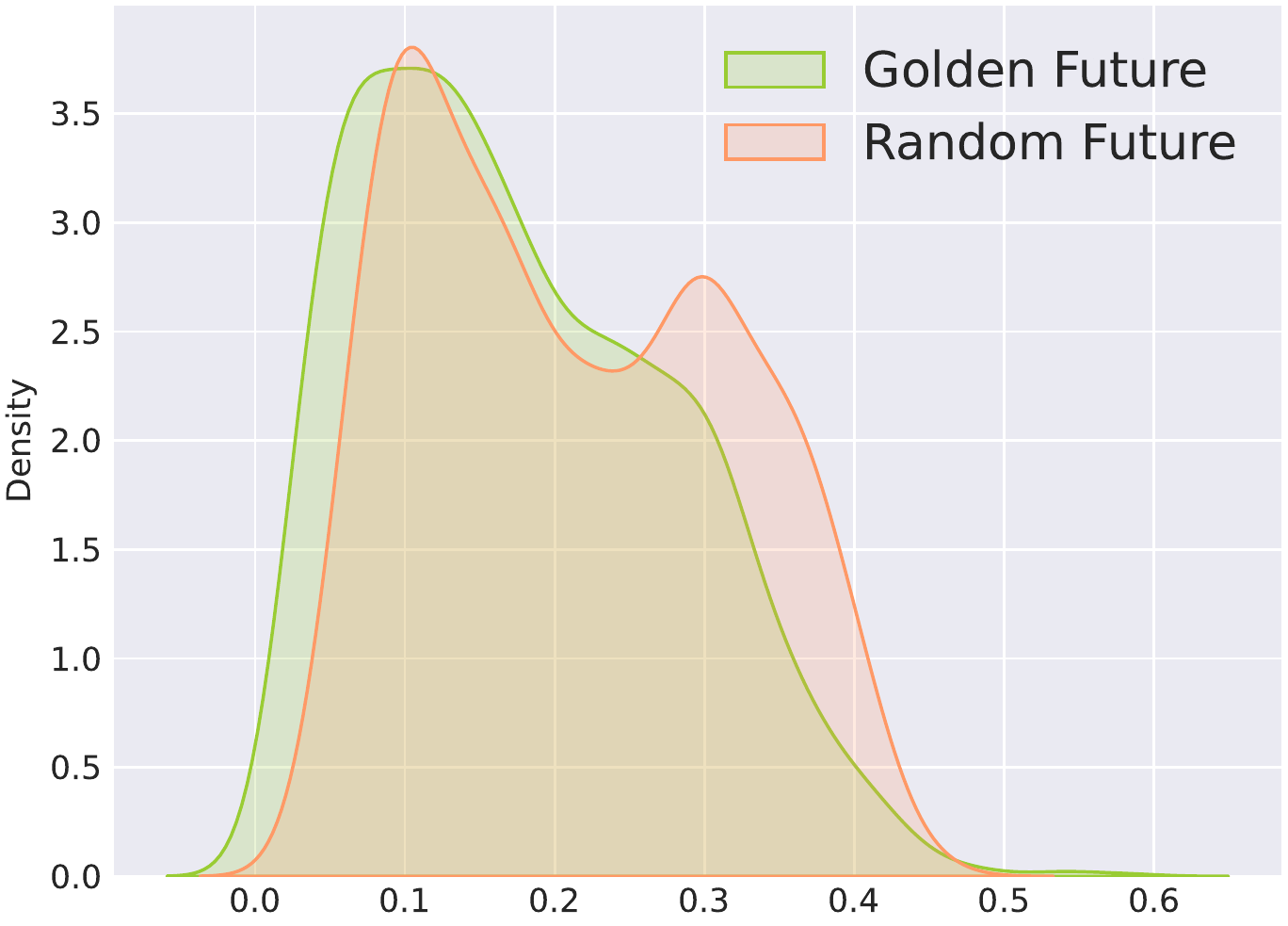}
    }
    \caption{Distance distribution curves of golden and random future. The X-axis denotes the MSE distance of  representations between the dialogue history and the concatenation of history and golden or random response. The Y-axis denotes the ratio.}
    \label{fig:curve}
    \end{adjustbox}
\end{figure}

To understand whether our FutureTOD can capture future knowledge, we perform a qualitative analysis to exhibit the capability of predicting future information in Figure \ref{fig:curve}. For each dialogue history, we combine its golden response with 99 randomly sampled responses. Then we compute the mean square error (MSE) distance between the representations of the dialogue history and the concatenation of history and response using a pre-trained FutureTOD model. For these randomly sampled responses, we report the average distance. Figure \ref{fig:curve} displays the distance distribution curves of golden and random future in the test set. The area under the shadow represents the ability of the model to predict the future. We find FutureTOD obtains similar representations corresponding to the golden future response. We also compute the average distance of all the test dialogues. We observe FutureTOD gets 1.449 of golden responses, smaller than 1.503 of random responses on MWOZ. Similar results are shown on DSTC2. They prove the effectiveness of FutureTOD capturing future knowledge.

\begin{figure}[t]
    \centering
    \begin{adjustbox}{minipage=\linewidth,scale=1.0}
    \subfigure[MWOZ]{
        \includegraphics[width=0.48\textwidth]{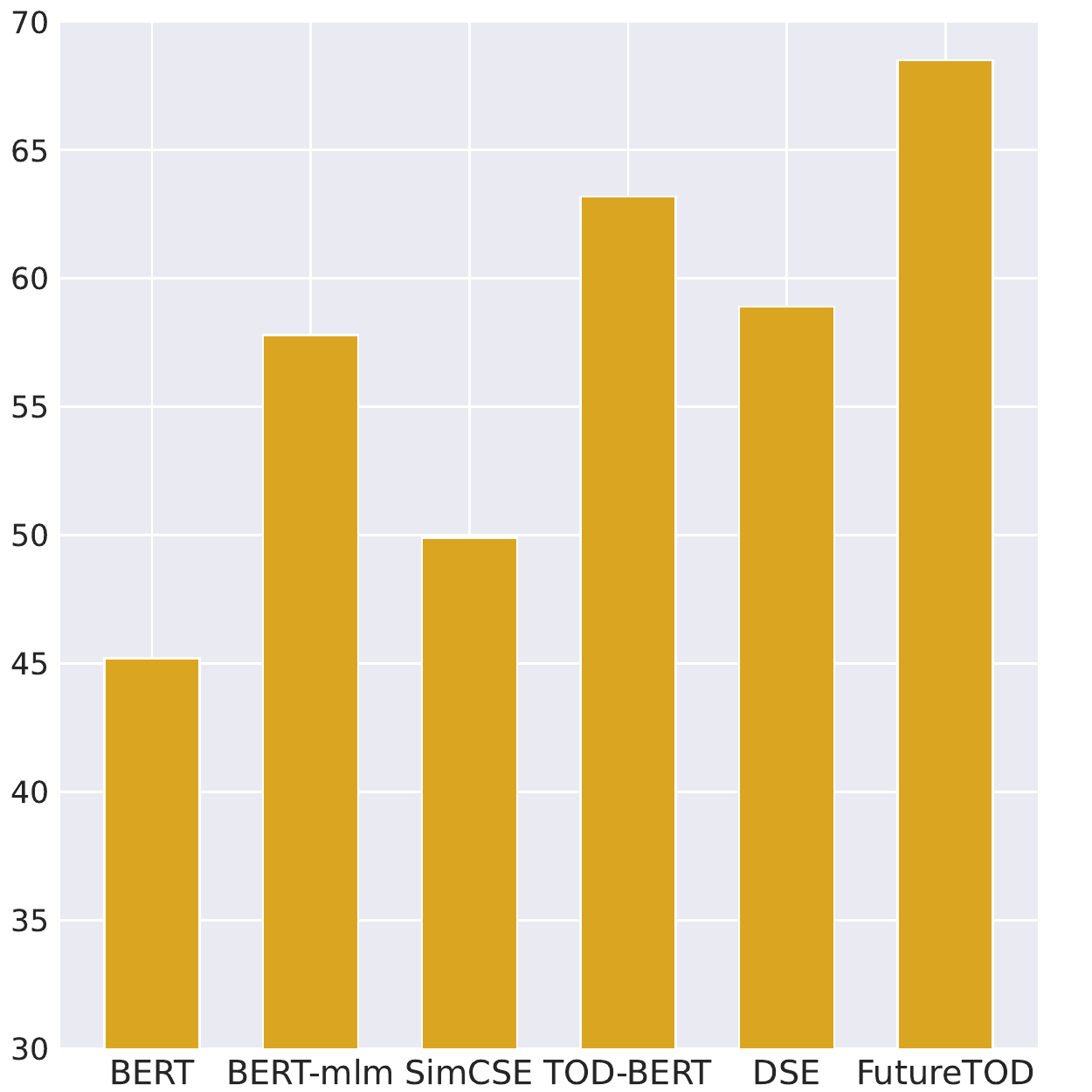}
    }
    \hspace{-0.5cm}
    \subfigure[DSTC2]{
        \includegraphics[width=0.48\textwidth]{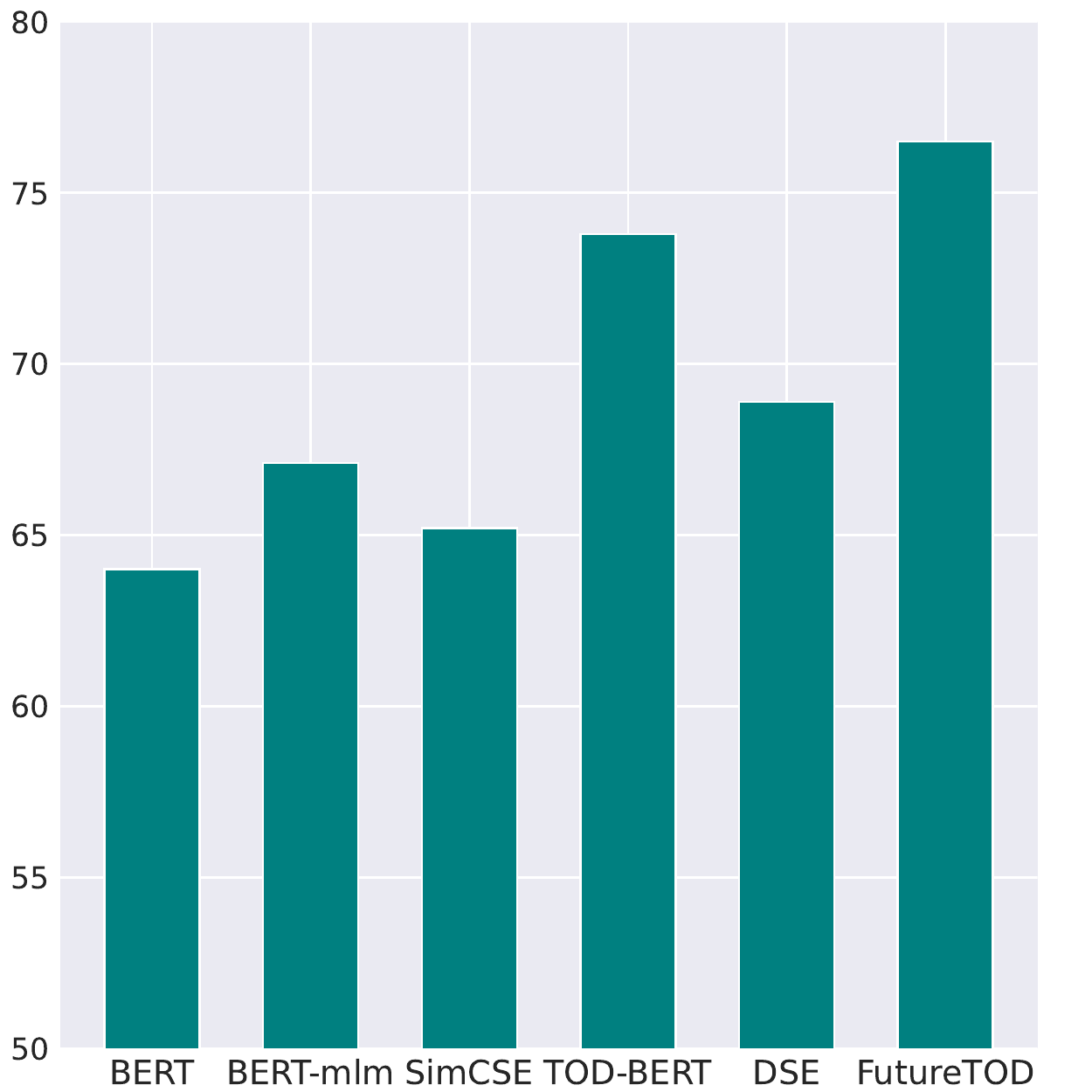}
    }
    \caption{The ratio of the test dialogue history where its distance between history and (history, golden response) is smaller than the one between history and (history, random response). Larger numbers denote better results.}
    \label{fig:ratio}
    \end{adjustbox}
\end{figure}

Besides, we compare different pre-trained models in predicting future information in Figure \ref{fig:ratio}. For each dialogue history in the test set, we compute the MSE distances between representations of dialogue history with/without golden or random responses. We assume the distances of golden responses are smaller than those of random responses. Therefore, we display the ratio of the test dialogue history where its distance of golden response is smaller than one of random response. As Figure \ref{fig:ratio} shows, we find FutureTOD obtains the highest ratio than the others, demonstrating the stronger capability of capturing future knowledge.

\subsection{Learning Process}
\begin{figure}[t]
    \centering
    \begin{adjustbox}{minipage=\linewidth,scale=1.0}
    \subfigure[Training curves]{
        \includegraphics[width=0.48\textwidth]{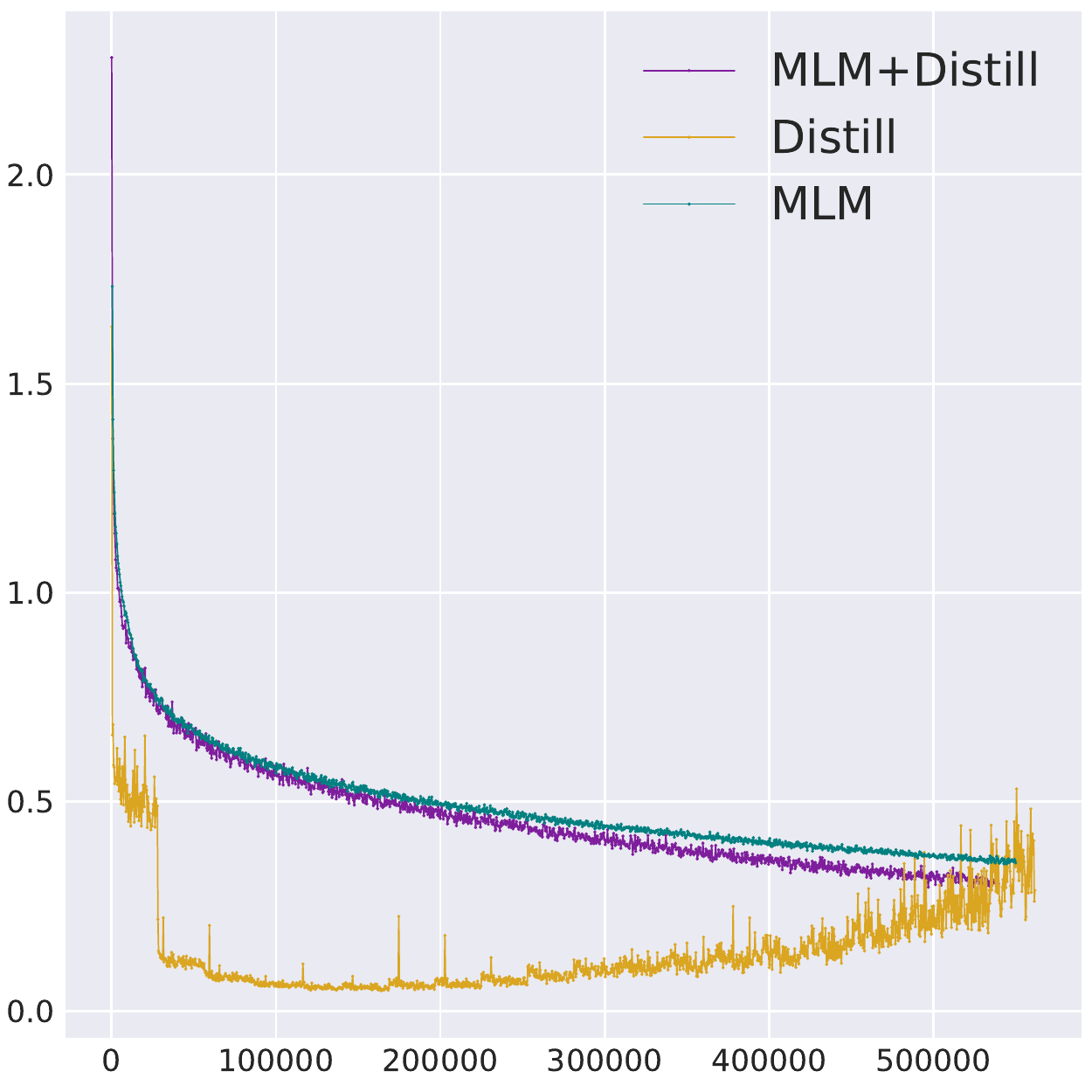}
    }
    \hspace{-0.5cm}
    \subfigure[Evaluation curves]{
        \includegraphics[width=0.48\textwidth]{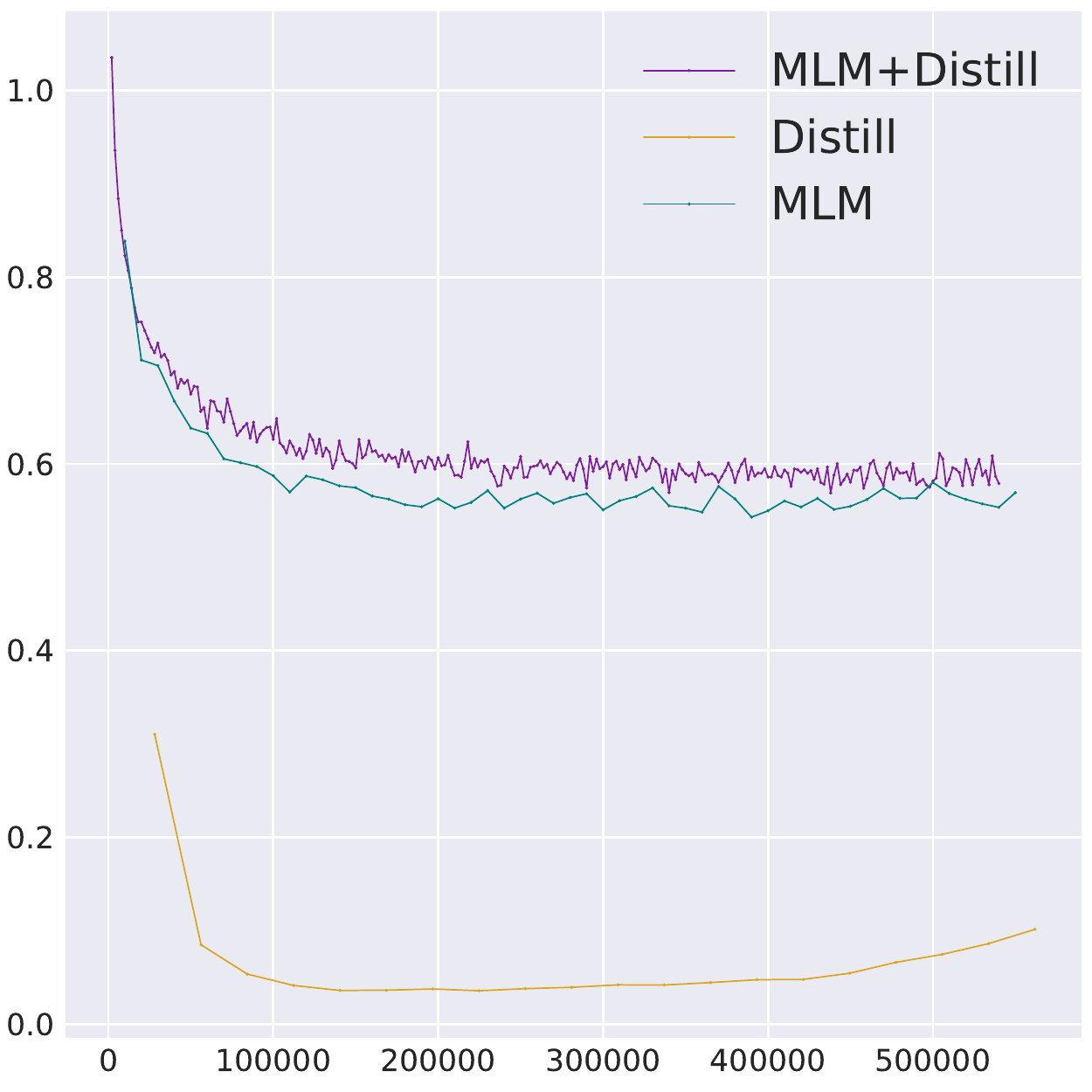}
    }
    \caption{Training and evaluation curves of different pre-training objectives. We scale up MLM loss by 50 times to display the three curves in the same figure.}
    \label{fig:train} 
    \end{adjustbox}
\end{figure}

Figure \ref{fig:train} displays the training and evaluation learning curves in the pre-training stage. We show three pre-training objectives: MLM, Distill, and MLM+Distill(FutureTOD). We find that only Distill loss leads to an unstable learning process and can't converge. We argue that adding random masks to the input sequence of the student model makes the architecture asymmetric between the student and teacher models, which is beneficial to preventing collapse. We also observe that adding another projection layer to the teacher model \cite{grill2020bootstrap} or momentum updating \cite{He2020MomentumCF} can't bring further improvements.

\section{Related Work}
\textbf{Self-Supervised Learning} Self-supervised learning (SSL) has been a very active area of research in CV, NLP, and speech. Contrastive methods \cite{chen2020simple,He2020MomentumCF} in computer vision achieve huge success in ImageNet. Further, \citet{Wu2020TODBERTPN,gao-etal-2021-simcse,Zhou2022LearningDR} in NLP introduce contrastive methods to unsupervised sentence or dialogue representation learning. 
However, these methods suffer from large batch size \cite{He2020MomentumCF}, easy negatives \cite{Wang2021UnderstandingTB}, and false negatives \cite{Huynh2022BoostingCS}. Besides, carefully designing appropriate augmentation methods \cite{Fang2020CERTCS,gao-etal-2021-simcse} is also challenging, especially in NLP. Another line of SSL is masked image/language/speech modeling. The most prominent model is BERT \cite{devlin-etal-2019-bert} which randomly masks some of the input tokens to recover from the remaining input. Vision methods follow similar ideas and predict visual tokens \cite{Dong2021PeCoPC} or input pixels \cite{He2022MaskedAA}. \citet{grill2020bootstrap,Baevski2022data2vecAG} use a momentum encoder to bridge the gap between different augmentation or masked views. Different from these works, we use future utterances to distill knowledge to the representation of the previous dialogue context without any augmentation.

\noindent\textbf{Dialogue Pre-trained Language Models} \citet{Zhang2020DIALOGPTL} adopts the pre-trained GPT-2 model \cite{Radford2019LanguageMA} on Reddit data to perform open-domain dialogue response generation. \citet{gao-etal-2021-simcse, Wu2020TODBERTPN,Zhou2022LearningDR} adopt contrastive learning to learn text or TOD dialogue representations. They use Dropout \cite{Srivastava2014DropoutAS} augmentation, context-response pair, and consecutive utterances to construct positive pairs, respectively. \citet{Henderson2020ConveRTEA,Liu2021DialogueCSEDC} use the similar idea to learn dialogue representations mainly for dialogue retrieval or response selection. Apart from these unsupervised methods, \citet{Zhou2022LearningDR,He2022SPACE2TS} use labeled dialogue data to perform supervised or semi-supervised pre-training. They usually use dialogue acts or dialogue NLI labels \cite{Williams2018ABC}. Since we focus on unsupervised pre-training in this paper, we don't compare these models and leave it to future work.

\section{Conclusion}
We propose a novel dialogue pre-training model, FutureTOD, which distills future knowledge to dialogue representations. Instead of existing contrastive works, we employ a simple self-training framework to learn from each other and dismiss the requirements of contrastive pairs. We perform comprehensive experiments on various task-oriented dialogue tasks, including intent classification, out-of-domain detection, dialogue state tracking, dialogue act prediction, and response selection. FutureTOD significantly outperforms TOD-BERT, DSE, and other strong baselines in all the scenarios. FutureTOD is of excellent performance and easy-to-deploy without modifying any model architecture.

\section*{Acknowledgements}
We thank all anonymous reviewers for their helpful comments and suggestions. We are also grateful to the track organizers for their valuable work. This work was partially supported by National Key R\&D Program of China No. 2019YFF0303300 and Subject II No. 2019YFF0303302, DOCOMO Beijing Communications Laboratories Co., Ltd, MoE-CMCC "Artifical Intelligence" Project No. MCM20190701. Jingang Wang is funded by Beijing Nova Program(Grant NO. 20220484098)

\section*{Limitations}
Although FutureTOD achieves significant improvements over existing baselines, there are some directions to explore for future work: (1) In this paper, FutureTOD doesn't use any data augmentation strategies to enhance representations. We believe existing augmentation methods will benefit further improving performance. (2) We design a simple technique of constructing the teacher. More complicated methods should be considered, such as multi-teacher and large teacher. (3) FutureTOD in this paper cares about dialogue understanding tasks like intent detection, dialogue state tracking, etc. We hope to extend the similar idea to the generative dialogue pre-trained models and larger TOD corpus. Besides, exploiting limited dialogue labels is also valuable to explore.

\section*{Ethics Statement}
The datasets used in this paper are all public and have been checked before use to not include any information that names or uniquely identifies individual people or offensive content. However, since the datasets come from the Internet, potential bias may still be introduced. This paper does not contain any data collection or release, so there are no privacy issues. Our model is pre-trained on GPU, which may cause an environmental impact. This paper does not involve human annotation or research with human subjects.

\bibliography{anthology,custom}
\bibliographystyle{acl_natbib}

\appendix

\section{Pre-training Data statistics}
\label{data}

We use the corpus collected by \citet{Wu2020TODBERTPN}, including 9 publicly available task-oriented datasets: MetaLWOZ \cite{Lee2019MultiDomainTD}, Schema \cite{Rastogi2020TowardsSM}, Taskmaster \cite{Byrne2019Taskmaster1TA}, MWOZ \cite{Budzianowski2018MultiWOZA}, MSR-E2E \cite{Li2018MicrosoftDC}, SMD \cite{Eric2017KeyValueRN}, Frames \cite{Asri2017FramesAC}, WOZ \cite{Mrksic2017NeuralBT}, CamRest676 \cite{RojasBarahona2017ANE}. The full statistics in Table \ref{tb:train_dataset}. These existing datasets are open-source and have no ethical concerns.


\section{Finetuning Details}




\label{sec:fintunedetails}

For BERT-mlm and TOD-BERT, we use the results reported by TOD-BERT \cite{Wu2020TODBERTPN} directly. We use the same hyperparameters for all the downstream tasks except the batch size and learning rate. We finetune all downstream tasks for 50 epochs with an early-stopped strategy evaluated on the validation set every 50 steps with patience set to 10. We respectively set batch size to 8, 25, 16 and 100 for intent recognition, dialogue state tracking, dialogue act prediction, and response selection. We choose the best learning rate from \{2e-5, 5e-5, 7e-5, 1e-4, 2e-4\} using grid search. We used the last layer's hidden states of the pre-trained model for downstream tasks. We also experimented with using hidden states from all layers, but find no significant change in performance.

\section{Only the Future}
\label{sec:only_the_future}

We use a student model to encode the context and a teacher model to encode both the context and the future in our method. We also conducted experiments using the teacher model without the context, but only with the future. However, as shown in Table \ref{abla_tea_encode}, the latter model did not perform well. For example, in response selection, the top-1 accuracy decreased from 58.4\% to 56.3\%, and the top-3 accuracy decreased from 72.6\% to 70.6\%. In dialogue act prediction, the micro-F1  decreased from 92.0\% to 90.9\%, and the macro-F1 decreased from 81.9\% to 81.3\%. We analyzed that this is due to the model collapse caused by directly aligning context and response without negative samples like TOD-BERT.

\section{Different Representation Methods}
\label{sec:mean_pool}
By default, we use the [CLS] token's representation as the utterance representation. To explore the impact of different utterance representation methods, we compare [CLS] token representations with the mean pooling of all the token representations. Table \ref{repre_method} shows that our FutureTOD model achieves comparable performance using both [CLS] and mean pooling. Both methods outperform the baselines. For instance, the FutureTOD(AVG) model achieves 87.0\% accuracy for the intent recognition task, while FutureTOD(CLS) achieves 87.2\%. These results surpass the 86.6\% accuracy achieved by TOD-BERT(CLS), demonstrating the robustness of our model across different representation methods.

\begin{table}[t]
\centering
\resizebox{1.0\linewidth}{!}{
\begin{tabular}{l|c|c|c|c}
\hline
\textbf{Name} & \textbf{\# Dialogue} & \textbf{\# Utterance} & \textbf{Avg. Turn} & \textbf{\# Domain} \\ \hline
MetaLWOZ & 37,884 & 432,036 & 11.4 & 47 \\ \hline
Schema & 22,825 & 463,284 & 20.3 & 17 \\ \hline
Taskmaster & 13,215 & 303,066 & 22.9 & 6 \\ \hline
MWOZ & 10,420 & 71,410 & 6.9 & 7 \\ \hline
MSR-E2E & 10,087 & 74,686 & 7.4 & 3 \\ \hline
SMD & 3,031 & 15,928 & 5.3 & 3 \\ \hline
Frames & 1,369 & 19,986 & 14.6 & 3 \\ \hline
WOZ & 1,200 & 5,012 & 4.2 & 1 \\ \hline
CamRest676 & 676 & 2,744 & 4.1 & 1 \\ \hline
\end{tabular}
}
\caption{Data statistics for our pre-training task-oriented dialogue datasets.}
\label{tb:train_dataset}
\end{table}
\begin{table}[t]
\centering

\begin{tabular}{c|c|cc}
\hline
\multirow{2}{*}{\textbf{Task}}                                                      & \multirow{2}{*}{\textbf{Metric}} & \multicolumn{2}{c}{\textbf{Method}} \\ \cline{3-4} 
                                                                                    &                                  & C $\leftrightarrow$ F              & C $\leftrightarrow$ C+F            \\ \hline
\multirow{2}{*}{\begin{tabular}[c]{@{}c@{}}Dialogue Act \\ Prediction\end{tabular}} & micro-F1                         & 90.9\%             & 92.0\%             \\
                                                                                    & macro-F1                         & 81.3\%             & 81.9\%             \\ \hline
\multirow{2}{*}{\begin{tabular}[c]{@{}c@{}}Response \\ Selection\end{tabular}}      & 1-to-100                         & 56.3\%             & 58.4\%             \\
                                                                                    & 3-to-100                         & 70.6\%             & 72.6\%             \\ \hline
\end{tabular}

\caption{Ablation of the Teacher Input. We report the results of dialogue act prediction on MWOZ and response selection on DSTC2. C $\leftrightarrow$ C+F denotes the teacher model that encodes both the context and the future(our default setting). C $\leftrightarrow$ F denotes the teacher model that encodes only the future, without the context.}
\label{abla_tea_encode}
\end{table}
\begin{table*}
\centering

\begin{tabular}{c|c|ccc}
\hline
\multirow{2}{*}{\textbf{Task}}                                                                        & \multirow{2}{*}{\textbf{Metric}} & \multicolumn{3}{c}{\textbf{Model}}                                             \\ \cline{3-5} 
                                                                                             &                         & \textbf{TOD-BERT(CLS)} & \textbf{FutureTOD(AVG)} & \textbf{FutureTOD(CLS)} \\ \hline
\multirow{4}{*}{\begin{tabular}[c]{@{}c@{}}Intent \\ Recognition\end{tabular}}        & Acc(all)                & 86.6\%            & 87.0\%                  & \textbf{87.2}\%                  \\
                                                                                             & Acc(in)                 & \textbf{96.2}\%            & 95.5\%                  & 96.0\%                  \\
                                                                                             & Acc(out)                & 89.9\%            & \textbf{90.2}\%                  & 90.0\%                  \\
                                                                                             & Recall(out)             & 43.6\%            & \textbf{48.8}\%                  & 47.6\%                  \\ \hline
\multirow{2}{*}{\begin{tabular}[c]{@{}c@{}}Dialogue State \\ Tracking\end{tabular}} & Joint Acc               & 48.0\%            & 50.1\%                  & \textbf{50.4}\%                  \\
                                                                                             & Slot Acc                & 96.9\%            & \textbf{97.1}\%                  & \textbf{97.1}\%                  \\ \hline
\multirow{2}{*}{\begin{tabular}[c]{@{}c@{}}Dialogue Act \\ Prediction\end{tabular}} & micro-F1                & 93.8\%            & \textbf{95.1}\%                  & 94.6\%                  \\
                                                                                             & macro-F1                & 41.3\%            & \textbf{45.9}\%                  & 44.6\%                  \\ \hline
\multirow{2}{*}{\begin{tabular}[c]{@{}c@{}}Response \\ Selection\end{tabular}}      & 1-to-100                & 56.8\%            & 57.7\%                  & \textbf{58.4}\%                  \\
                                                                                             & 3-to-100                & 70.6\%            & 72.5\%                  & \textbf{72.6}\%                  \\ \hline
\end{tabular}

\caption{Ablation study of different representation methods. We report the results of intent recognition on OOS, DST on MWOZ, dialogue act prediction on DSTC2, and response selection on DSCT2. TOD-BERT (CLS) and FutureTOD (CLS) denote using CLS token representation as the utterance representation. FutureTOD (AVG) denotes using the mean pooling of all tokens within the utterance as the utterance representation.}
\label{repre_method}
\end{table*}

\end{document}